\documentclass{fairmeta}

\usepackage{amsmath,amssymb,amsthm,bm}
\usepackage{graphicx,booktabs,multirow,wrapfig,float}
\usepackage{microtype}
\usepackage[round,authoryear]{natbib}
\usepackage{hyperref}
\hypersetup{colorlinks=true,linkcolor=metablue,citecolor=metablue,urlcolor=metablue}
\usepackage{causalalign_metapaired}

\usepackage{amsmath,amsfonts,bm}

\def\eqref#1{equation~\ref{#1}}

\def\1{\bm{1}}

\DeclareMathAlphabet{\mathsfit}{\encodingdefault}{\sfdefault}{m}{sl}
\SetMathAlphabet{\mathsfit}{bold}{\encodingdefault}{\sfdefault}{bx}{n}

\title{CausalAlign: Causal Alignment of Video Generation Models for Consistent Ego-Motion}

\author[1,*]{Kai Zhu}
\author[1,*]{Yintianrun Zhang}
\author[1]{Shuwei Jiang}
\author[1]{Ruolin Zuo}
\author[2]{Yixin Chen}
\author[3]{Duomin Wang}
\author[4]{Miao Liu}
\author[5]{Yin Li}
\author[1]{Daquan Zhou}
\author[1,\dagger]{Yiwu Zhong}

\affiliation[1]{Peking University}
\affiliation[2]{Beijing Institute for General Artificial Intelligence}
\affiliation[3]{NVIDIA}
\affiliation[4]{Tsinghua University}
\affiliation[5]{University of Wisconsin--Madison}

\contribution[*]{Equal contribution}
\contribution[\dagger]{Corresponding author}

\abstract{Diffusion-based video generation models (VGMs) can produce visually plausible videos, yet their camera ego-motion often drifts over long horizons. Camera ego-motion determines how the viewpoint changes between frames and is essential for maintaining consistent geometry and trajectories.
To address this problem, we propose \textbf{CausalAlign}, a causal alignment framework with two components.
First, we diagnose how precise ego motion is represented, used, and retained within VGMs using linear probes and causal interventions, revealing a rise-and-fall pattern during denoising.
Second, we strengthen ego-motion state in VGM by aligning the representations at the identified layers to a 3D foundation model through delta-token space. This space encodes the changes between consecutive frames, allowing the alignment to focus on ego-motion dynamics rather than static scene content.
With Wan2.2 fine-tuned on DL3DV and RealEstate10K, CausalAlign can largely improve motion accuracy, geometric consistency, and temporal smoothness, while preserving visual quality. For example, it improves geometric consistency by more than \textbf{20\%} relative to the base model.
Moreover, with the fine-tuned VGM, CausalAlign improves the success rate of world action models (WAMs) by \textbf{10 points} on mobile manipulation tasks of RoboCasa365 benchmark.
Overall, CausalAlign improves ego-motion consistency in diffusion-based VGMs and yields noticeable gains in embodied models.}

\metadata[Code]{\href{https://github.com/PKU-VaLuE-Lab/CausalAlign}{github.com/PKU-VaLuE-Lab/CausalAlign}}
\metadata[Project Page]{\href{https://pku-value-lab.github.io/CausalAlign-homepage/}{pku-value-lab.github.io/CausalAlign-homepage/}}

\begin{document}

\maketitle
\setlength{\parindent}{0pt}
\setlength{\parskip}{0.35em}

\section{Introduction}

Recent advances in diffusion-based video generation models (VGMs) have led to remarkable models such as Wan2.2~\citep{wan2025} and Cosmos3~\citep{agarwal2026cosmos}. 
Trained on Internet-scale collections of 2D videos, these models can synthesize visually compelling content~\citep{huang2024vbench,huang2025vbenchpp}. 
However, trained to model 2D appearance and local motion statistics, standard VGMs struggle to capture 3D camera trajectories and to control viewpoint changes consistently across frames~\citep{wu2026geometry,kupyn2025epipolar}. We refer to these cross-frame viewpoint changes as camera ego-motion.

\begin{figure}[t]
    \centering
    \includegraphics[
        width=\textwidth,
        height=0.48\textheight,
        keepaspectratio
    ]{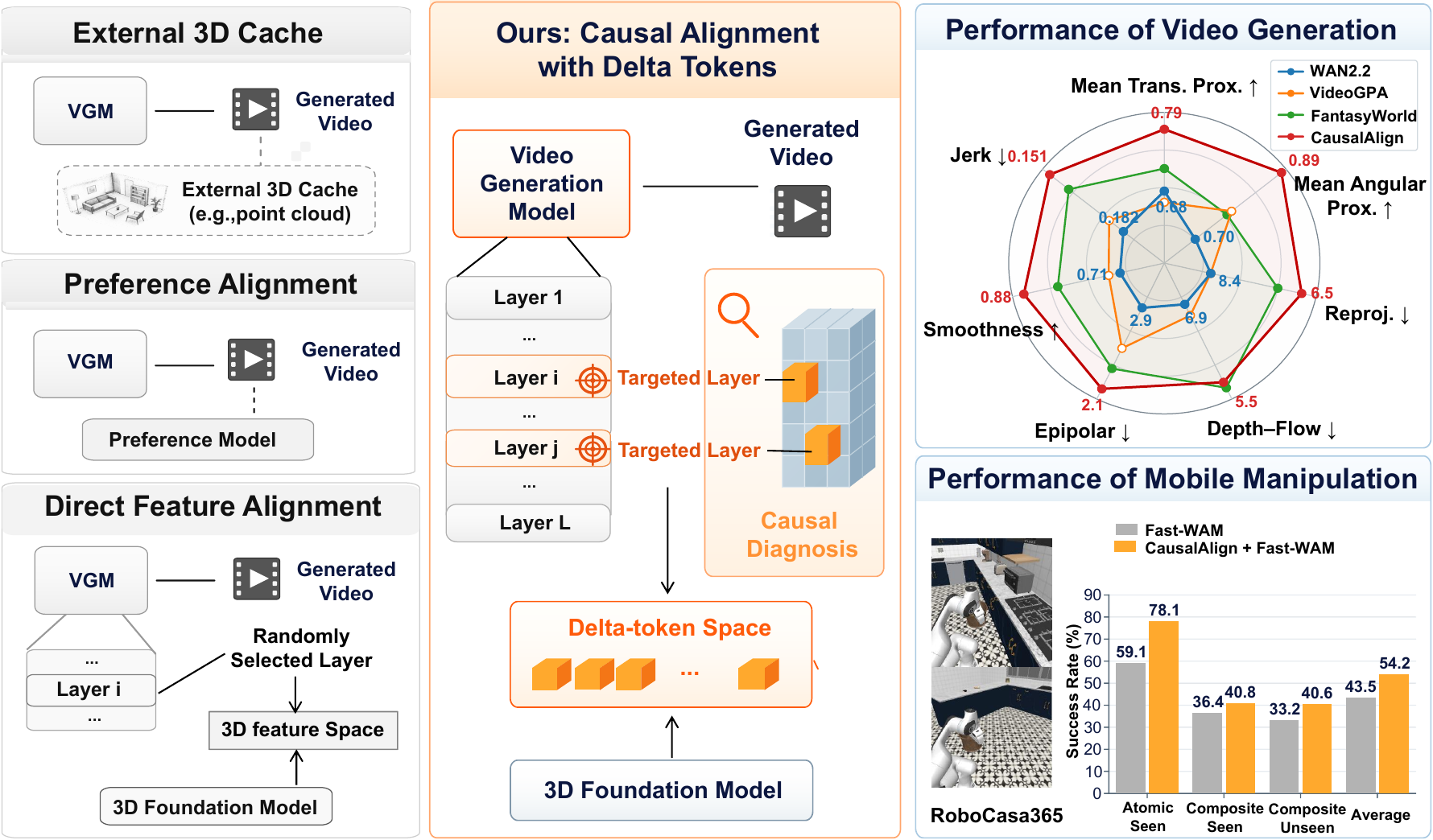}
    
\caption{
    \textbf{Overview of CausalAlign.}
    Left: Existing approaches rely on external 3D caches that may propagate errors, optimize output-level preferences without directly supervising ego-motion states, or align 3D features at randomly selected layers.
    Middle: Our CausalAlign first identifies the targeted layers through causal diagnosis and then strengthens ego-motion states through delta-token alignment.
    Right: CausalAlign consistently improves camera ego-motion consistency in video generation and largely boosts downstream mobile manipulation performance.
}

    \label{fig:comparison}
\end{figure}

As shown in Figure ~\ref{fig:comparison} (left panel), prior work has attempted to address this challenge by introducing explicit 3D priors or preference alignment~\citep{wang2024motionctrl,ren2025gen3c,zhang2025worldconsistent,bahmani2025ac3d}, yet generating videos with geometrically consistent camera trajectories remains an open problem. 
At the same time, recent work shows that VGMs can implicitly estimate camera poses from their internal representations, even when their generated trajectories are imperfect~\citep{bahmani2025ac3d}. This raises a more fundamental question: {{\textbf{how is precise ego-motion represented, used, and retained within VGMs?}}} 
This question becomes increasingly important when VGMs are integrated into world action models (WAMs) for robotic mobile manipulation: visually plausible yet geometrically inconsistent predictions of the future can induce systematic errors in downstream action prediction.

As the first step, we perform analysis for understanding the ego-motion information within VGMs. 
Inspired by prior works~\citep{alain2016understanding,meng2022locating}, we adopt linear probing to measure the internal representations of VGMs and causal intervention to reveal the mechanisms of VGMs. 
Specifically, for linear probing, we define ``decodability'' as the amount of ego-motion information. Meanwhile, for causal intervention, we define ``causal effect'' as the extent to which the interventions on VGM can change the generation. 
Our analysis results demonstrate that both measurements exhibit the same trend, that is, increasing first and then declining, across both denoising timesteps and denoising Transformer layers. 
These results suggest that ego-motion state emerges as the denoising progresses, however, fades away at the late timesteps and the last few Transformer layers at each step.
Therefore, ego-motion is already learned and represented in VGMs, increasingly used at intermediate stages, yet not well retained at later stages. We refer to this progressive pattern as \textbf{``geometric state forgetting''}.


Motivated by this diagnosis, our goal is to preserve and strengthen the ego-motion state within VGMs without compromising the quality of visual generation.
To this end, we propose \textbf{CausalAlign}, a framework with two components: causal diagnosis and delta-token alignment, as shown in Figure~\ref{fig:comparison} (middle panel).
First, causal diagnosis leverages linear probing and causal intervention to identify where ego-motion is represented, used, and forgotten within a VGM.
Second, delta-token alignment supervises the internal ego-motion state at the positions identified as forgetting.
There are two key challenges in our alignment: \textit{how to represent ego-motion with the hidden states in VGMs}, and \textit{how to provide meaningful geometry supervision for such representations}.
For the first challenge, we adopt the idea of delta tokens, which was initially designed for improving the efficiency of the world model~\citep{kerssies2026frame}. We adapt this idea to measure the ego-motion via the difference between two consecutive frames. Delta tokens serve as a natural carrier for motion information while reducing redundant static scene content. 
For the second challenge, we construct the delta-token space by leveraging the pretrained 3D foundation model. These 3D models are capable of estimating camera poses, depth, and 3D structure, and thus can provide high-quality geometric supervision. 
A learnable projector then maps consecutive hidden-state transitions into this teacher delta-token space, explicitly supervising ego-motion state of the VGM.
By aligning these projected tokens with the corresponding teacher delta tokens at the identified layers, CausalAlign transfers geometric knowledge to the locations where ego-motion information starts to fade.

With our CausalAlign method, we fine-tune the VGM on the widely-adopted datasets, DL3DV~\citep{ling2024dl3dv} and RealEstate10K~\citep{zhou2018stereo}. 
To evaluate the effects of CausalAlign on VGMs, we establish a comprehensive evaluation protocol with three aspects: visual quality, ego-motion consistency, and geometric state retention. 
Visual quality is measured with VBench-I2V~\citep{huang2025vbenchpp} and complementary metrics.
Ego-motion consistency is evaluated through motion magnitude, geometric consistency, and temporal dynamics, while geometric-state retention is assessed through probe decodability and causal intervention effects.
CausalAlign preserves visual quality while improving camera ego-motion consistency, as shown in Figure~\ref{fig:comparison} (right panel).
For example, it reduces reprojection error by \textbf{22.6\%} and improves trajectory smoothness by \textbf{23.9\%} relative to the base model.
It also alters the decodability and causal-effect patterns from \emph{rise-and-fall} to \emph{rise-and-maintain}, supporting the improved retention of ego-motion state.
Moreover, we observe that the improved ego-motion state can facilitate downstream applications, such as WAMs for embodied tasks. Specifically, our fine-tuned VGM raises success rate on RoboCasa365~\citep{nasiriany2026robocasa365} from 43.5\% to \textbf{54.2\%}. These results validate our method with causal diagnosis and delta-token alignment.

Our contributions are summarized as follows: 
(1) We identify a pattern called \textbf{``geometric state forgetting''} which reveals the ego-motion state in diffusion-based VGMs. Through linear probing and causal intervention, our analysis reveals that ego-motion information is progressively represented and used during the denoising process, yet not retained at the late stages of denoising. 
(2) We propose \textbf{CausalAlign}, a causal alignment method that aims to preserve ego-motion state at the VGM layers identified by our causal analysis. Our technical innovation lies in adapting the idea of delta tokens to strengthen ego-motion consistency via 3D foundation models.
(3) Extensive experiments and comprehensive metrics validate the benefits of our method. It largely improves ego-motion consistency, better preserves ego-motion state during denoising, and offers remarkable gains for WAM in robotic tasks.

\section{Related Work}
\label{sec:related_work}

\noindent\textbf{Video diffusion and world action models.}
Video diffusion extends image denoising to joint spatiotemporal generation~\citep{ho2022video}, while open backbones such as Wan have broadened its use~\citep{wan2025}. 
Unlike image generation, VGMs require preserving temporal correspondences and scene geometry across multiple frames~\citep{huang2024vbench,huang2025vbenchpp,ren2025gen3c,zhang2025worldconsistent}. 
These priors increasingly serve as action-conditioned simulators or policies: Genie and IRASim learn controllable future dynamics, while VideoVLA, DreamZero, and LingBot-VA couple observations with executable actions; Fast-WAM shows that video co-training can improve action representations without test-time video decoding~\citep{bruce2024genie,zhu2025irasim,shen2026videovla,ye2026worldaction,li2026lingbotva,yuan2026fastwam}. 
Therefore, ego-motion consistency is internally related to WAM and erroneous camera motion can corrupt the predicted actions. 
Rather than proposing another WAM, CausalAlign restores and strengthens the ego-motion state of the VGM to facilitate WAM performance. 
Our method elicits the knowledge that has been learned during VGM pretraining, providing an alternative route for stronger WAMs. 

\vspace{-1em}
\paragraph{Feed-forward 3D foundation models.}
Feed-forward 3D reconstruction has progressed from point-map predictors such as DUSt3R and VGGT~\citep{wang2024dust3r,wang2025vggt} to streaming
models with temporal state. CUT3R recurrently maintains a metric-scale scene representation~\citep{wang2025cut3r}, while LingBot-Map organizes streaming context into anchor, pose-reference, and trajectory memories and demonstrates improved trajectory and pose accuracy over VGGT~\citep{chen2026lingbotmap}. These models provide camera and scene representations that can serve as supervision for video generation. We thus adopt LingBot-Map as the teacher model and study how its ego-motion state can be transferred to a video generator. 

\vspace{-1em}
\paragraph{Geometry-aware alignment and post-training.}
As summarized in Figure~\ref{fig:comparison}, existing geometry-aware adaptation methods can be categorized into three types. 
External 3D cache methods such as GEN3C and Voyager use point cloud or world memories for controllable camera motion, yet require additional geometric memory and may propagate cache errors~\citep{ren2025gen3c,huang2025voyager}. 
Preference alignment methods such as VideoGPA and Epipolar-DPO supervise generated videos through preference or epipolar feedback, without directly improving the VGM's internal geometric state~\citep{du2026videogpa,kupyn2025epipolar}. 
Direct alignment methods, including JOG3R, Geometry Forcing, and FantasyWorld, transfer geometric features into the generator but do not explicitly target at effective timestep and layer locations~\citep{huang2025jog3r,wu2026geometry,dai2026fantasyworld}. 
Complementary WAM methods leverage 3D, 4D, or action-grounded supervision to improve action representations without requiring auxiliary teachers at deployment~\citep{li2026wam4d,zhang2026mecowam,yang2026fourdwam,fu2026fourdwamdriving,qiu2026agra}. 
CausalAlign instead improves the internal ego-motion state of VGMs through delta-token space from 3D foundation models, at the locations identified by causal diagnosis. 

\begin{figure*}[t]
    \centering
    \includegraphics[
        width=\textwidth,
        height=0.50\textheight,
        keepaspectratio
    ]{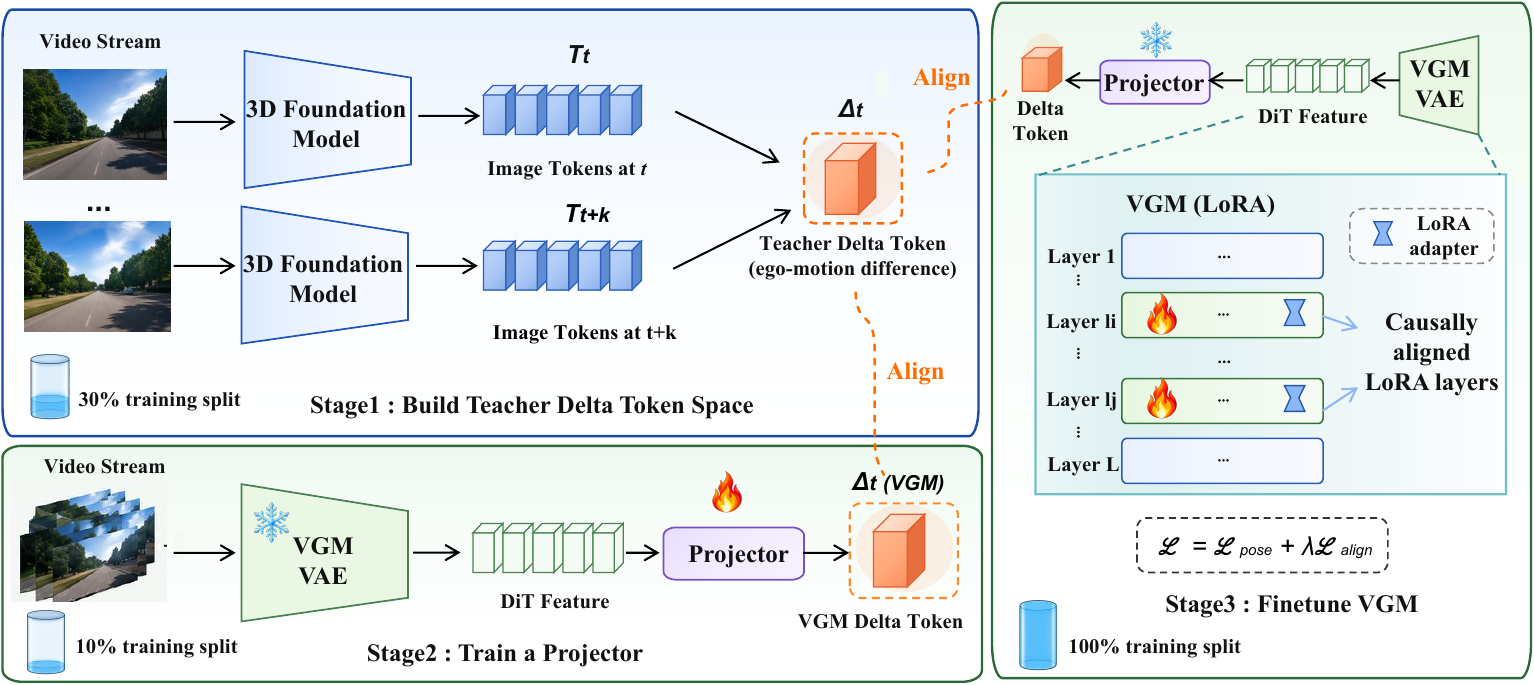}
    \vspace{-1.5em}
    \caption{\textbf{Three stages of CausalAlign training.} CausalAlign transfers ego-motion knowledge from a 3D teacher to the VGM through three-stage training. In Stage 1, consecutive features from a streaming 3D foundation model are encoded into a teacher delta-token space that captures inter-frame geometric transitions. In Stage 2, a projector maps VGM DiT features into the same delta-token space and is frozen after training, using a small subset of the training split. In Stage 3, with full training split, LoRA adapters are optimized only at the identified VGM layers using delta tokens, while the projector remains frozen.}

\label{fig:method}
\end{figure*}

\section{Method}
\vspace{-0.5em}
\subsection{Method Overview}

CausalAlign combines causal diagnosis and delta-token alignment to preserve ego-motion representations in diffusion-based VGMs. 
We first conduct causal diagnosis to identify the layers and the denoising timesteps where ego-motion information fades.
Guided by the causal diagnosis, we align the VGM using delta tokens in three stages. First, we learn delta tokens from a frozen 3D foundation model. Second, we train projectors to map the temporal transitions of VGM features into this space. Finally, we use these tokens to supervise LoRA updates at the identified layers and denoising timesteps, while keeping the original VGM backbone, teacher, and projectors frozen. 
In this way, CausalAlign strengthens the geometric knowledge encoded in the VGM without introducing additional computation during video generation.

\subsection{Causal Diagnosis of Ego-Motion States}
\label{sec:causal_targeting}

\noindent\textbf{Decodability.}
Let $\mathbf{h}_t^{\ell,s}$ denote the spatially aggregated VGM state at
frame $t$, layer $\ell$, and noise-level index $s$. A shared operator
$\Gamma$ constructs the consecutive-frame transition feature
\[
\mathbf{u}_{t,\ell,s}
=
\Gamma\!\left(
\mathbf{h}_t^{\ell,s},
\mathbf{h}_{t+1}^{\ell,s}
\right).
\]
A linear probe predicts the relative camera motion
\begin{equation}
\widehat{\boldsymbol{\xi}}_{t,\ell,s}
=
\mathbf{W}_{\ell,s}\mathbf{u}_{t,\ell,s}
+
\mathbf{b}_{\ell,s},
\qquad
\boldsymbol{\xi}_t
=
\operatorname{Log}
\!\left(
\mathbf{T}_t^{-1}\mathbf{T}_{t+1}
\right),
\label{eq:linear_probe}
\end{equation}
where $\mathbf{T}_t\in\mathrm{SE}(3)$ is the ground-truth
camera-to-world pose and $\operatorname{Log}$ maps the relative pose to
a six-dimensional motion vector. The held-out score
$D_{\ell,s}$ measures how well ego-motion is decodable at location
$(\ell,s)$; decodability alone does not establish causal use
~\citep{meng2022locating,geiger2021causal}.

\noindent\textbf{Causal intervention and target selection.}
At $c=(\ell,s)$, we keep the prompt, initial noise, and random seed fixed
and intervene on the temporal hidden state using
\[
\mathcal{A}
=
\{
\textsc{Mask},
\textsc{Shuffle},
\textsc{Replace},
\textsc{Inject}
\}.
\]
These interventions remove temporal information, disrupt its order,
replace it with another trajectory, or insert a pose-matched donor,
respectively.

For diagnostic clip $d$, let
$\mathbf{T}^{\mathrm{ref},d}_{1:T}$ be its ground-truth
camera-to-world trajectory from the corresponding real video. A frozen
pose estimator produces $\widehat{\mathbf{T}}(Y)$ from a generated video
$Y$, and we define
\begin{equation}
E_{\mathrm{traj}}(Y;d)
=
\mathcal{E}_{\mathrm{traj}}
\!\left(
\widehat{\mathbf{T}}(Y),
\mathbf{T}^{\mathrm{ref},d}_{1:T}
\right).
\label{eq:trajectory_error}
\end{equation}
The reference trajectory is used only for evaluation and is not provided
as a generation condition. For intervention $a\in\mathcal{A}$, the
norm-matched random-control-adjusted effect is
\begin{equation}
\overline{\mathcal{C}}_{c}^{(a)}
=
\eta_a\,
\mathbb{E}_{d}
\left[
E_{\mathrm{traj}}
\!\left(Y_c^{(a)};d\right)
-
E_{\mathrm{traj}}
\!\left(Y_c^{(\mathrm{rand},a)};d\right)
\right],
\label{eq:adjusted_causal_effect}
\end{equation}
where $\eta_a=+1$ for the disruptive interventions and
$\eta_a=-1$ for \textsc{Inject}, so that larger values indicate stronger
causal effects. The aggregate causal score is
\begin{equation}
C_{\ell,s}
=
\frac{1}{|\mathcal{A}|}
\sum_{a\in\mathcal{A}}
\overline{\mathcal{C}}_{(\ell,s)}^{(a)}.
\label{eq:aggregate_causal_score}
\end{equation}

Averaging $D_{\ell,s}$ and $C_{\ell,s}$ over the complementary axis
gives layer profiles ($A=\mathrm{L}$) and timestep profiles
($A=\mathrm{S}$). We combine them as
\begin{equation}
J^A
=
\frac{1}{2}
\left[
\mathcal{N}(D^A)
+
\mathcal{N}(C^A)
\right],
\qquad
A\in\{\mathrm{L},\mathrm{S}\},
\label{eq:joint_diagnostic_score}
\end{equation}
where $\mathcal{N}$ independently min--max normalizes each profile.
The target sets $\mathcal{L}_{\mathrm{tar}}$ and
$\mathcal{S}_{\mathrm{tar}}$ are selected at locations with the largest
positive decline rates of $J^A$. The corresponding normalized timestep
declines provide the weights $\omega_{\ell,s}$ used by the alignment
losses and shared across selected layers.

\subsection{Delta-Token Alignment}
\label{sec:delta_alignment}

\noindent\textbf{Stage 1: learning teacher delta tokens.}
For a $T$-frame training video $\mathbf{x}_{1:T}$, the frozen 3D foundation model $\Phi_{\mathrm{T}}$ extracts geometric states $\mathbf{z}_t^{\mathrm{T}}=\Phi_{\mathrm{T}}(\mathbf{x}_{1:t})$. A trainable encoder $E_\Delta$ maps states to a delta token:
\begin{equation}
    \boldsymbol{\delta}_t^{\mathrm{T}}
    =E_\Delta(\mathbf{z}_t^{\mathrm{T}},\mathbf{z}_{t+1}^{\mathrm{T}})
    \in\mathbb{R}^{d_\Delta},
    \label{eq:teacher_delta_token}
\end{equation}
where $d_\Delta$ is the token dimension; superscripts $\mathrm{T}$ and $\mathrm{V}$ denote teacher and VGM representations. A transition decoder reconstructs the next state from the current state and token, a temporal predictor forecasts the next token, and a camera decoder $D_{\mathrm{cam}}$ predicts relative pose. These auxiliary tasks encourage the token to retain transition information rather than merely compressing the teacher state. Denoting their predictions by $\widehat{\mathbf{z}}_{t+1}^{\mathrm{T}}$, $\widehat{\boldsymbol{\delta}}_{t+1}^{\mathrm{T}}$, and $\widehat{\Delta\mathbf{T}}_t^{\mathrm{T}}$, we train the encoder and auxiliary modules with
\begin{equation}
\begin{aligned}
    \mathcal{L}_{\mathrm{T}}
    ={}&\mathbb{E}_t\!\left[
    \|\widehat{\mathbf{z}}_{t+1}^{\mathrm{T}}-\mathbf{z}_{t+1}^{\mathrm{T}}\|_2^2
    +\lambda_{\mathrm{temp}}
    \|\widehat{\boldsymbol{\delta}}_{t+1}^{\mathrm{T}}-\boldsymbol{\delta}_{t+1}^{\mathrm{T}}\|_2^2
    \right]\\
    &+\lambda_{\mathrm{pose}}^{\mathrm{T}}\mathbb{E}_t\!\left[
    \ell_{\mathrm{pose}}(\widehat{\Delta\mathbf{T}}_t^{\mathrm{T}},\Delta\mathbf{T}_t)
    \right].
\end{aligned}
\label{eq:teacher_delta_loss}
\end{equation}
Here, $\Delta\mathbf{T}_t=\mathbf{T}_t^{-1}\mathbf{T}_{t+1}$ is the ground-truth relative pose used only for supervision, and $\ell_{\mathrm{pose}}$ combines geodesic rotation distance with normalized SmoothL1 translation error. The expectation averages valid training-frame positions; $\lambda_{\mathrm{temp}}$ and $\lambda_{\mathrm{pose}}^{\mathrm{T}}$ weight temporal and pose supervision. These objectives preserve transition structure and anchor tokens to camera motion. We then freeze the encoder and auxiliary modules; decoder architectures and pose-loss details are provided in the appendix.

\noindent\textbf{Stage 2: learning VGM-side projectors.}
With the VGM and teacher modules frozen, layer-specific projectors $Q_\ell$ map corresponding VGM  features to teacher space tokens $\boldsymbol{\delta}_{t,\ell,s}^{\mathrm{V}}=Q_\ell(\mathbf{u}_{t,\ell,s})$. We minimize
\begin{equation}
    \mathcal{L}_{\mathrm{map}}
    =\sum_{\ell\in\mathcal{L}_{\mathrm{tar}}}
     \sum_{s\in\mathcal{S}_{\mathrm{tar}}}\omega_{\ell,s}
     \mathbb{E}_t\!\left[
     \|\boldsymbol{\delta}_{t,\ell,s}^{\mathrm{V}}-\boldsymbol{\delta}_t^{\mathrm{T}}\|_2^2
     \right],
    \label{eq:projector_mapping_loss}
\end{equation}
where $\omega_{\ell,s}$ weights selected timesteps by their normalized positive diagnostic-score decline rates, shared across selected layers. The projectors are frozen after training.

\noindent\textbf{Stage 3: adapting the VGM.}
We update LoRA adapters~\citep{hu2022lora} at selected layers. 
For an adapted transition $\widetilde{\mathbf{u}}_{t,\ell,s}$, frozen projector produces
$\widetilde{\boldsymbol{\delta}}_{t,\ell,s}^{\mathrm{V}}
=Q_\ell(\widetilde{\mathbf{u}}_{t,\ell,s})$.
We use two complementary objectives:
\begin{equation}
    \mathcal{L}_{\Delta}
    =
    \sum_{\ell\in\mathcal{L}_{\mathrm{tar}}}
    \sum_{s\in\mathcal{S}_{\mathrm{tar}}}
    \omega_{\ell,s}\,
    \mathbb{E}_t\!\left[
    \left\|
    \widetilde{\boldsymbol{\delta}}_{t,\ell,s}^{\mathrm{V}}
    -
    \boldsymbol{\delta}_t^{\mathrm{T}}
    \right\|_2^2
    \right],
    \label{eq:delta_alignment_loss}
\end{equation}
which aligns the adapted VGM transitions with the teacher delta tokens, and
\begin{equation}
    \mathcal{L}_{\mathrm{pose}}
    =
    \sum_{\ell\in\mathcal{L}_{\mathrm{tar}}}
    \sum_{s\in\mathcal{S}_{\mathrm{tar}}}
    \omega_{\ell,s}\,
    \mathbb{E}_t\!\left[
    \ell_{\mathrm{pose}}\!\left(
    D_{\mathrm{cam}}\!\left(
    \widetilde{\boldsymbol{\delta}}_{t,\ell,s}^{\mathrm{V}}
    \right),
    \Delta\mathbf{T}_t
    \right)
    \right],
    \label{eq:pose_alignment_loss}
\end{equation}
which explicitly supervises the predicted relative camera motion. 
The final adaptation objective is
\begin{equation}
    \mathcal{L}_{\mathrm{CausalAlign}}
    =
    \mathcal{L}_{\mathrm{pose}}
    +
    \lambda_{\Delta}\mathcal{L}_{\Delta},
    \label{eq:CausalAlign_objective}
\end{equation}
where $\lambda_{\Delta}$ controls the strength of token alignment. 
Gradients pass through the frozen projectors and camera decoder to the LoRA adapters, while all other parameters remain frozen. 
The teacher is used only to provide training targets, and no external geometric cache is required during generation.

\subsection{Inference}
\label{sec:causalalign_inference}

At inference, CausalAlign follows the same video generation procedure as the original Wan2.2~\citep{wan2025}.
Given an initial frame and a text prompt, the adapted VGM generates a video through the standard denoising and decoding process.
All auxiliary modules used during training, including the 3D foundation model, are no longer needed.
Therefore, CausalAlign improves ego-motion consistency without requiring additional inputs or geometric computation during video generation.

\subsection{Application to World Action Models}
\label{sec:CausalAlign_wam_method}

Consistent ego-motion is important for mobile manipulation, where robots predict actions while their viewpoints change.
We thus investigate whether improving the ego-motion state of VGMs can also benefit action prediction of world action models.
Specifically, we initialize the video expert of Fast-WAM~\citep{yuan2026fastwam} with our adapted Wan2.2 model.
For the action expert, we follow Fast-WAM and transform the parameters of the original Wan model to initialize ActionDiT.
We train both experts using the original objectives for flow matching:
\begin{equation}
    \mathcal{L}_{\mathrm{WAM}}
    =\mathcal{L}_{\mathrm{FM}}^v+\mathcal{L}_{\mathrm{FM}}^a,
    \label{eq:CausalAlign_wam_loss}
\end{equation}
where $v$ and $a$ denote the video and action experts, respectively.
The policy receives synchronized images from three camera views and predicts actions starting from the current observation.
Future video is used as a training target, while at inference, the policy predicts executable actions from the observed history without generating future video~\citep{yuan2026fastwam}.

\begin{figure*}[t]
    \centering
    \includegraphics[width=\textwidth]{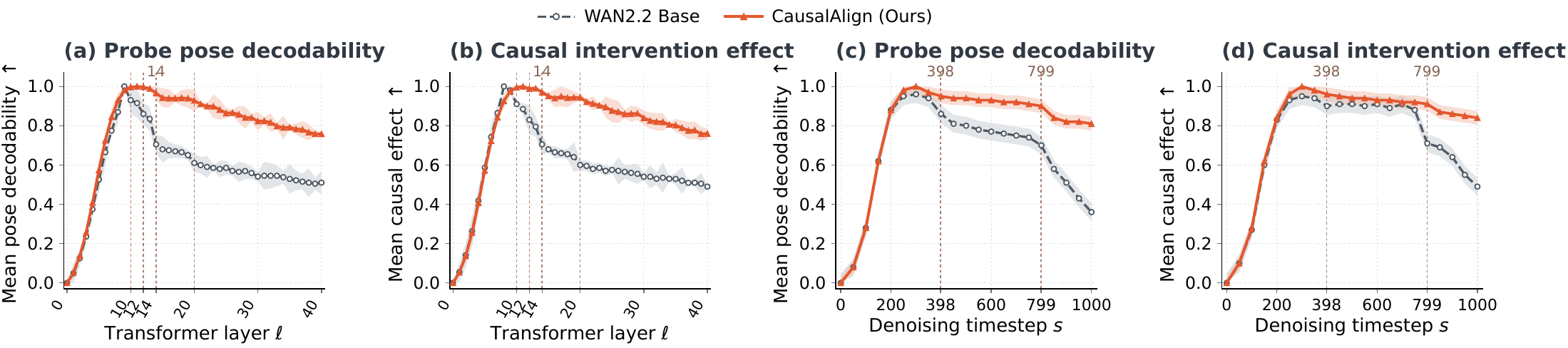}

    \caption{%
  \textbf{Ego-motion diagnosis and retention.}
  We compare base model Wan2.2 and our method CausalAlign through probe decodability and causal intervention effects across layers (sub-figure (a) \& (b): results are averaged over timesteps) and across denoising timesteps (sub-figure (c) \& (d): results are averaged over layers). 
  Dashed vertical lines represent the selected alignment locations and shading shows the 10th--90th percentiles over the averaged dimension. 
  Our method CausalAlign can raise both signals from rise-and-fall to rise-and-maintain pattern, indicating improved ego-motion retention.
}
    \label{fig:causal_diagnosis}
\end{figure*}

\section{Experiment}
We first conduct causal diagnosis to identify where ego-motion information fades and examine whether CausalAlign improves its retention.
We then fine-tune the VGM with delta-token supervision at these locations and evaluate its ego-motion consistency, visual quality, and the benefits for embodied tasks.
Finally, we conduct ablation studies to validate our design choices.

\subsection{Causal Diagnosis and Target Selection}
\label{sec:exp_causal_target}

\noindent\textbf{Setup.}
We freeze Wan2.2~\citep{wan2025} and diagnose the hidden states across Transformer layers and across denoising steps.
At each location $c=(\ell,s)$, a linear readout predicts relative camera motion, yielding a decodability score $\mathcal{D}_{\ell,s}$.
We also apply the causal interventions introduced in Section~\ref{sec:causal_targeting} and measure their effects on the generated camera trajectory, yielding a causal effect score $\mathcal{C}_{\ell,s}$.
Figure~\ref{fig:causal_diagnosis} summarizes the resulting layer--timestep profiles.

\noindent\textbf{Diagnosis and target selection.}
Both geometric decodability and causal effect in the original Wan2.2 exhibit a consistent rise-and-fall pattern.
Their increase at intermediate locations indicates that the VGM forms and uses an ego-motion state, while their subsequent decline suggests that this state is not retained during later denoising.
By applying the target selection rule from Section~\ref{sec:causal_targeting}, we select layers 14, 10, 12, and 20 in descending order of their layer-wise decline rates.
Layer 14 is the primary target, while layers 10, 12, and 20 are auxiliary targets.
The same rule applies to denoising timesteps $s\in\{398,799\}$.
Both selected timesteps are used at all selected layers, with weights proportional to their positive temporal decline rates.

\noindent\textbf{The benefits of our method.} After alignment, CausalAlign retains both probe decodability and causal effects at later denoising stages, changing the original rise-and-fall pattern into a rise-and-maintain pattern, as shown by the orange curve in Figure~\ref{fig:causal_diagnosis}.
The sustained causal effects indicate that ego-motion information remains influential during generation, rather than merely becoming more decodable.

\subsection{Experiments on the Comprehensive Evaluation}
\label{sec:comprehensive_protocol}


\noindent\textbf{Setup.}
We apply CausalAlign to Wan2.2-14B-I2V~\citep{wan2025} and train it on datasets DL3DV~\citep{ling2024dl3dv} and RealEstate10K~\citep{zhou2018stereo}.
We choose these datasets since they provide camera pose annotations and cover diverse scenes and camera movements, making them suitable for supervising and evaluating ego-motion consistency.
For each dataset, 90\% of the data is used for training and 10\% of the data is for evaluation.
As in Figure~\ref{fig:method}, we first learn delta tokens from the geometric features of a frozen 3D foundation model using 30\% of the training data.
We then train projectors to map features from the video generator into this delta-token space using 10\% of the training data.
Finally, we fine-tune the generator with LoRA at the identified layers and denoising steps using the full training split. 
The 3D teacher and the projectors are needed only during training.
For evaluation, we use the same video length, resolution, frame rate, sampling settings, and random seeds across methods, and average results over multiple samples for each input.
Further details are provided in Appendix~\ref{app:training_details}.

\noindent\textbf{Baselines.}
We compare CausalAlign with the base model Wan2.2 and three groups of methods for improving geometric consistency.
\emph{External 3D cache methods}, including GEN3C~\citep{ren2025gen3c} and Voyager~\citep{huang2025voyager}, control camera motion using explicit trajectories and 3D caches.
We follow their standard inference procedures by providing the reference camera trajectory and constructing a new 3D cache from the available input frames, without using future frames.
\emph{Preference alignment methods}, including VideoGPA~\citep{du2026videogpa} and Epipolar-DPO~\citep{kupyn2025epipolar}, improve generated videos through preference or epipolar feedback.
\emph{Direct feature alignment methods}, including JOG3R~\citep{huang2025jog3r}, Geometry Forcing~\citep{wu2026geometry}, and FantasyWorld~\citep{dai2026fantasyworld}, transfer geometric features into the generator.
For Wan2.2, CausalAlign, and methods without explicit camera control, we provide only the first frame and a text prompt.

\noindent\textbf{Metrics.}
We use the same frozen estimators~\citep{wang2025vggt} to recover camera poses, depth, optical flow, and visual features from every generated video.
Our evaluation covers three complementary aspects.
\emph{Motion Magnitude} measures how closely the mean translational and angular speeds match the corresponding means from real validation videos (Appendix~\ref{app:metric_definitions}).
\emph{Geometric Consistency} measures reprojection error~\citep{triggs1999bundle}, consistency between depth and optical flow in static regions~\citep{luo2019every}, and epipolar error~\citep{lee2020geometric}.
\emph{Temporal Dynamics} measures trajectory smoothness and jerk~\citep{hogan2009sensitivity,flash1985coordination}, capturing continuity of motion and abrupt changes in acceleration.
Since the first frame and a text prompt can admit several valid future camera trajectories, we measure intrinsic consistency rather than error against a single reference trajectory (Appendix~\ref{app:training_details}).

\begin{table*}[t]
\caption{%
  \textbf{Camera ego-motion consistency.}
  We evaluate videos generated from the DL3DV and RealEstate10K evaluation splits with the same estimators and generation settings. Methods without camera control receive the first frame and text prompt; methods that require control also receive camera trajectories and 3D caches. Bold indicates the best result for each metric. CausalAlign improves all seven metrics over Wan2.2 and leads on six of seven metrics.
}
\label{tab:ego_motion_comparison}

\centering
\setlength{\tabcolsep}{2pt}

\resizebox{\linewidth}{!}{%
\begin{tabular}{@{}lccccccc@{}}
\toprule
\multicolumn{1}{c}{\bfseries Method}
& \multicolumn{2}{c}{\bfseries Motion Magnitude}
& \multicolumn{3}{c}{\bfseries Geometric Consistency}
& \multicolumn{2}{c}{\bfseries Temporal Dynamics} \\
\cmidrule(lr){2-3}
\cmidrule(lr){4-6}
\cmidrule(lr){7-8}
&
\shortstack{Mean Trans. Prox.\\$\uparrow$}
&
\shortstack{Mean Angular Prox.\\$\uparrow$}
&
\shortstack{Reproj.\\(px)$\downarrow$}
&
\shortstack{Depth--Flow\\(px)$\downarrow$}
&
\shortstack{Epipolar\\(px)$\downarrow$}
&
\shortstack{Smoothness\\$\uparrow$}
&
\shortstack{Jerk\\$\downarrow$} \\
\midrule

\multicolumn{8}{l}{\emph{(a) Base Model}} \\
Wan2.2
& 0.68 & 0.70 & 8.4 & 6.9 & 2.9 & 0.71 & 0.182 \\

\midrule
\multicolumn{8}{l}{\emph{(b) External 3D Cache}} \\
GEN3C
& 0.63 & 0.75 & 7.8 & 6.5 & 2.8 & 0.71 & 0.175 \\
Voyager
& 0.69 & 0.73 & 7.1 & 6.8 & 2.8 & 0.73 & 0.179 \\

\midrule
\multicolumn{8}{l}{\emph{(c) Preference Alignment}} \\
VideoGPA
& 0.66 & 0.78 & 8.4 & 6.7 & 2.5 & 0.73 & 0.176 \\
Epipolar-DPO
& 0.68 & 0.72 & 7.8 & 5.9 & 2.2 & 0.78 & 0.162 \\

\midrule
\multicolumn{8}{l}{\emph{(d) Direct Feature Alignment}} \\
JOG3R
& 0.70 & 0.75 & 7.2 & 5.9 & 2.5 & 0.79 & 0.158 \\
Geometry Forcing
& 0.69 & 0.78 & 6.9 & 5.7 & 2.4 & 0.80 & 0.164 \\
FantasyWorld
& 0.72 & 0.77 & 7.0 & \textbf{5.4} & 2.3 & 0.82 & 0.159 \\

\midrule
\multicolumn{8}{l}{\emph{(e) Causal Delta-Token Alignment}} \\
CausalAlign (Ours)
& \textbf{0.79}
& \textbf{0.89}
&  \textbf{6.5} 
&  5.5
& \textbf{2.1}
& \textbf{0.88}
& \textbf{0.151} \\

\bottomrule
\end{tabular}%
}
\end{table*}

\begin{figure*}[t]
    \centering
    \includegraphics[
        width=\textwidth,
        height=0.58\textheight,
        keepaspectratio
    ]{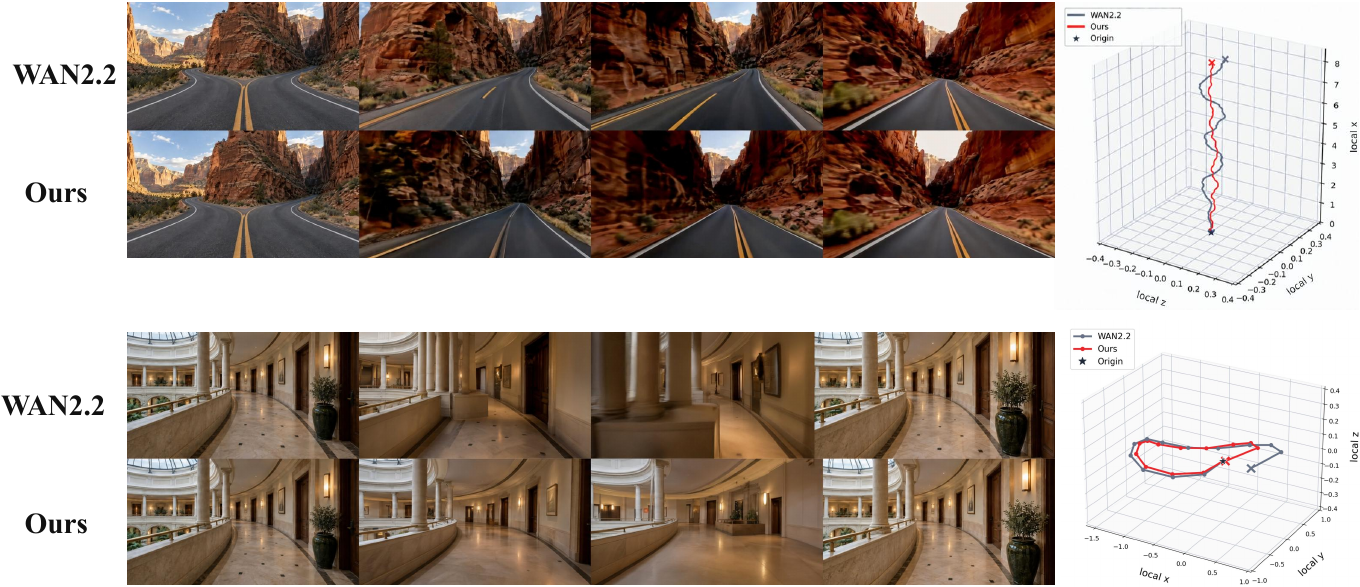}
    \caption{%
  \textbf{Qualitative comparison of ego-motion.}
  We compare frames and reconstructed camera trajectories from Wan2.2 and CausalAlign in two scenes. CausalAlign shows less trajectory jitter in the top example and less drift when the camera completes a loop in the bottom example.
}
    \label{fig:loop}
\end{figure*}

\noindent\textbf{Comparison with Baseline Methods.}
As shown in Table~\ref{tab:ego_motion_comparison}, CausalAlign improves all seven metrics over Wan2.2 and achieves the best results on six of seven metrics among the compared methods.
For example, it reduces reprojection error from 8.4 to 6.5 pixels (22.6\%), raises trajectory smoothness from 0.71 to 0.88 (23.9\%), and lowers jerk from 0.182 to 0.151.
Further, its proximity to real camera speeds improves for both translation (0.68 to 0.79) and rotation (0.70 to 0.89), while epipolar error decreases from 2.9 to 2.1 pixels.
FantasyWorld achieves a slightly lower error for consistency between depth and flow (5.4 versus 5.5 pixels), but CausalAlign leads on the other six metrics without requiring a camera trajectory.
These results suggest that causal alignment through delta tokens strengthens the generator's ability to produce consistent camera motion and scene geometry.
The examples in Figure~\ref{fig:loop} further show less trajectory jitter and drift when the camera completes a loop, compared with Wan2.2.

\paragraph{Visual Quality Preservation.}
Further, we examine whether stronger geometric supervision harms visual quality. Specifically, we evaluate CausalAlign and Wan2.2 on the same evaluation clips with identical prompts, sampling settings, and random seeds.
We report VBench-I2V quality, I2V score, and smoothness~\citep{huang2024vbench,huang2025vbenchpp}, together with FVD, CLIP, SSIM, and LPIPS~\citep{unterthiner2018towards,radford2021learning,wang2004image,zhang2018unreasonable}.
As shown in Table~\ref{tab:visual_preservation}, CausalAlign preserves visual quality comparable to Wan2.2: VBench quality increases slightly from 0.805 to 0.806, I2V score remains 0.958, and VBench smoothness rises from 0.983 to 0.985.
Further, CLIP and SSIM improve slightly, LPIPS remains unchanged, and FVD changes only marginally (182.4 to 182.7).
These results indicate that our method can improve ego-motion consistency without sacrificing visual fidelity.

\begin{table*}[t]
\caption{%
  \textbf{Visual quality preservation.}
  We evaluate Wan2.2 and CausalAlign using VBench-I2V (quality, I2V score, and smoothness) together with FVD, CLIP, SSIM, and LPIPS. Bold indicates the best result for each metric, including ties. CausalAlign preserves visual quality comparable to Wan2.2.
}
\label{tab:visual_preservation}

\setlength{\tabcolsep}{2.5pt}

\centering
\begin{tabular}{@{}c@{\hspace{0.08in}}c@{}}

\begin{tabular}[t]{@{}lcccc@{}}
\hline
\multicolumn{1}{c}{}
& \multicolumn{3}{c}{\bf VBench-I2V} \\
\cline{2-4}
&
\multicolumn{1}{c}{Quality$\uparrow$}
&
\multicolumn{1}{c}{I2V Score$\uparrow$}
&
\multicolumn{1}{c}{Smooth.$\uparrow$}
 \\
\hline
Wan2.2
& 0.805
& \bf 0.958
& 0.983 \\
CausalAlign
& \bf 0.806
& \bf 0.958
& \bf 0.985 \\
\hline
\end{tabular}

&

\begin{tabular}[t]{@{}lcccc@{}}
\hline
\multicolumn{1}{c}{\bf }
& \multicolumn{4}{c}{\bf Other Visual Metrics} \\
\cline{2-5}
&
\multicolumn{1}{c}{FVD$\downarrow$}
&
\multicolumn{1}{c}{CLIP$\uparrow$}
&
\multicolumn{1}{c}{SSIM$\uparrow$}
&
\multicolumn{1}{c}{LPIPS$\downarrow$} \\
\hline
Wan2.2
& \bf 182.4
& 0.311 
& 0.742
& \bf 0.183 \\
CausalAlign
& 182.7
& \bf 0.314
& \bf 0.744
& \bf 0.183 \\
\hline
\end{tabular}

\end{tabular}
\end{table*}

\paragraph{Application to World Action Models.}
Given that CausalAlign improves the retention of ego-motion information in VGMs, we wonder whether this benefit can transfer to the embodied models which are built on top of VGMs.
We initialize the video expert of Fast-WAM with our fine-tuned Wan2.2 and evaluate the resulting policy on RoboCasa365 Target100~\citep{nasiriany2026robocasa365}.
Following Fast-WAM, the policy is pretrained on Human300 and then trained with 500 target demonstrations per task. The Atomic Seen evaluation includes 18 tasks with 50 rollouts each.
After 50k steps of fine-tuning, the policy takes images from three cameras and the current proprioceptive state as input. It predicts 32 actions in ten denoising steps and executes eight actions before replanning at 20\,Hz.
As shown in Table~\ref{tab:cadet_wam}, CausalAlign + Fast-WAM raises average success from 43.5\% to 54.2\% (+10.7 percentage points) and success on Atomic Seen tasks from 59.1\% to 78.1\% (+19.0 points, 703/900 successful rollouts).
Further, success on Composite Seen tasks increases from 36.4\% to 40.8\%, and success on Composite Unseen tasks rises from 33.2\% to 40.6\%, suggesting that stronger ego-motion representations can benefit mobile manipulation.
Additional rollouts are shown in Appendix~\ref{sec:wam_vis}.

\begin{table}[t]
\centering
\small
\caption{%
  \textbf{Transfer to world action models.}
  We report success rates (\%) on the three RoboCasa365 Target100 task groups. Bold marks the best value in each column, including ties. CausalAlign improves Fast-WAM across all three groups, increasing average success from 43.5\% to 54.2\% and Atomic-Seen success by 19.0 percentage points.
}
\label{tab:cadet_wam}
\setlength{\tabcolsep}{4pt}
\begin{tabular*}{\linewidth}{@{\extracolsep{\fill}}lrrrr@{}}
\toprule
Method & \shortstack{Atomic Seen} & \shortstack{Composite Seen} & \shortstack{Composite Unseen} & Average \\
\midrule
GR00T-N1.5~\citep{nvidia20256} & 60.6 & 35.0 & 33.3 & 43.7 \\
Fast-WAM~\citep{yuan2026fastwam} & 59.1 & 36.4 & 33.2 & 43.5 \\
LingBot-VA~\citep{li2026lingbotva} & 63.5 & 37.3 & 32.1 & 45.1 \\
ABot-M0.5~\citep{chen2026abot} & 70.6 & \textbf{44.3} & \textbf{45.6} & \textbf{54.2} \\
\midrule
CausalAlign + Fast-WAM & \textbf{78.1} & 40.8 & 40.6 & \textbf{54.2} \\
\bottomrule
\end{tabular*}
\end{table}

\subsection{Ablation Studies}
\label{sec:cadet_ablations}

\begin{table*}[!t]
\centering\normalsize
\caption{%
  \textbf{Delta-token ablations.}
  We compare (a) semantic regions with matched-width PCA projections of equally supervised unpartitioned encoders and (b) 128D delta tokens with original 8192D/4096D LingBot-Map features. Best values within each panel are bold. Delta tokens lead all six metrics in both panels.
}
\label{tab:cadet_core}
\setlength{\tabcolsep}{2pt}
\begin{tabular*}{\textwidth}{@{\extracolsep{\fill}}lrrrrrr@{}}
\toprule
& \multicolumn{2}{c}{Ego motion} & \multicolumn{2}{c}{Geometry} & \multicolumn{2}{c}{Visual} \\
\cmidrule(lr){2-3}\cmidrule(lr){4-5}\cmidrule(lr){6-7}
Representation & Rotation ($^\circ$)$\downarrow$ & Direction ($^\circ$)$\downarrow$ & AbsRel$\downarrow$ & $\delta_1\uparrow$ & SSIM$\uparrow$ & LPIPS$\downarrow$ \\
\midrule
\multicolumn{7}{l}{\textit{(a) Matched-width partitions}} \\
Unpartitioned pair & 2.453 & 8.582 & 0.2426 & 0.7280 & 0.2924 & 0.08816 \\
Unpartitioned diff. & 2.837 & 10.392 & 0.2301 & 0.7390 & 0.2971 & 0.08754 \\
Delta regions & \textbf{2.244} & \textbf{7.565} & \textbf{0.2292} & \textbf{0.7457} & \textbf{0.3050} & \textbf{0.08643} \\
\midrule
\multicolumn{7}{l}{\textit{(b) Complete space versus original features}} \\
Original A (8192D) & 3.243 & 12.964 & 0.2698 & 0.6993 & 0.2838 & 0.08958 \\
Original B (4096D) & 4.276 & 13.451 & 0.2731 & 0.6947 & 0.2866 & 0.08958 \\
Delta (128D) & \textbf{2.530} & \textbf{8.231} & \textbf{0.2314} & \textbf{0.7422} & \textbf{0.3028} & \textbf{0.08676} \\
\bottomrule
\end{tabular*}
\end{table*}

\noindent\textbf{Semantic partitioning.}
We first investigate whether separating delta-token into motion, geometry, and visual information helps the model learn a more effective representation.
To this end, we compare our structured delta tokens, consisting of 64D ego-motion, 40D geometry, and 24D visual regions, with two unpartitioned 128D encoders.
All encoders receive the same supervision, and we use PCA to project the unpartitioned representations to the corresponding dimensions.
As shown in Table~\ref{tab:cadet_core}(a), our structured representation outperforms both alternatives across all six metrics.
Further, the cross-probes in Appendix~\ref{sec:cadet_partition_protocol} show that each region performs best on its corresponding task.
These results suggest that explicitly separating different types of information helps the model learn more effective representations of ego-motion.

\noindent\textbf{Complete-space comparison.}
We next examine whether learning delta tokens provides advantages over directly using the raw features of a pretrained 3D foundation model.
We compare our 128D delta tokens with two representations from LingBot-Map: the concatenated features of consecutive frames (8192D) and their difference (4096D).
Both original representations retain their full dimensions.
As shown in Table~\ref{tab:cadet_core}(b), our delta tokens outperform both alternatives across all six metrics, despite using substantially fewer dimensions.
For example, the rotation error decreases from 3.243 to 2.530 degrees compared with the original feature pair.
These results suggest that learning a compact representation of temporal changes is more effective than directly using the original pooled 3D features.


\section{Conclusion}
We introduce \textbf{CausalAlign}, a causal alignment framework for ego-motion consistent video generation. 
By linear probing and causal interventions, we show that video generation models already learned and used ego-motion state during denoising but did not retain it well at the late stages, a pattern we call ``geometric state forgetting''. 
CausalAlign targets at the locations where such forgetting is identified, and aligns ego-motion states with the delta-token space from a 3D foundation model. 
Extensive experiments demonstrate that CausalAlign consistently improves ego-motion consistency and temporal dynamics while preserving visual quality. It mitigates geometric state forgetting by changing the pattern of ego-motion state from rise-and-fall to rise-and-maintain. 
Further, our improved VGM largely raises the success rate of WAM on the mobile manipulation tasks of RoboCasa365. 
We hope this work can inspire the research on preserving geometric states of generative video models and the research on enhancing embodied models.

\bibliography{references}
\bibliographystyle{plainnat}

\clearpage\beginappendix
The appendix provides details needed to interpret and reproduce CausalAlign.
Appendix~\ref{app:detailed_method} specifies the causal diagnosis, target selection, and three-stage
delta-token alignment procedure, and provides additional evidence for
geometric-state forgetting. Appendix~\ref{app:extended_evaluation} defines the evaluation metrics and
normalization, clarifies the training and trajectory protocols, and extends
the experiments to text-to-video and camera-trajectory-conditioned generation.
Appendix~\ref{sec:cadet_extended_ablations} shows more ablation experiments to demonstrate the contributions of structured delta representations, update placement, temporal modulation, and auxiliary training objectives.

\section{Detailed Method Description}
\label{app:detailed_method}

\subsection{Causal Diagnosis and Target Selection}
\label{app:causal_diagnosis}

\paragraph{Diagnostic subset.}
We reserve a fixed diagnostic subset
$\mathcal{D}_{\mathrm{diag}}$ containing approximately $10\%$ of the complete
experimental data. 
The same subset is used to generate Fig.~\ref{fig:causal_diagnosis} and to
select the target layers and denoising timesteps. 
Target selection is performed before adaptation using only the measurements from
the frozen WAN2.2 base model~\citep{wan2025}. 
After adaptation, we re-evaluate the same subset only to visualize the change in
the geometric state dynamics. 
The diagnostic subset is not used for the held-out evaluation or downstream WAM
evaluation, and the selected target locations remain fixed in all subsequent
experiments.

\paragraph{Noise schedule and reindexed denoising progress.}
Let $r$ denote the native WAN2.2 scheduler timestep. Under the native
scheduler convention, denoising starts from a large $r$ with high noise and
proceeds toward a smaller $r$ with low noise. For diagnosis, we reverse this
ordering and use a reindexed denoising-progress coordinate $s$. Specifically,
$s=\pi(r)$ denotes the order-reversing mapping used in our implementation,
so smaller $s$ corresponds to an earlier, higher-noise state and larger $s$
corresponds to a later, lower-noise state. All diagnostic curves, decline-rate
computations, target selection, and alignment losses use this reindexed
coordinate $s$, rather than the native scheduler timestep $r$.

For each diagnostic real video, we inject the native scheduler noise
corresponding to $r=\pi^{-1}(s)$ and extract the VGM hidden states from the
resulting denoising computation. The probe features are therefore obtained
from noised real videos rather than from free-running generated videos. For
the intervention analysis, we instead perform free-running generation from
the same first frame, prompt, initial noise, and random seed, and intervene
at the selected layer--progress locations. We follow the native WAN2.2
expert routing and noise schedule; no additional routing rule is introduced
for diagnosis.

\paragraph{Ego-motion decodability.}
For each layer--timestep location $c=(\ell,s)$, a linear probe~\citep{alain2016understanding} predicts the
relative camera motion $\boldsymbol{\xi}_t$ from the VGM transition feature
$\mathbf{u}_{t,\ell,s}$. 
Let
\begin{equation}
e_{\ell,s}
=
\mathbb{E}_{t\sim\mathcal{D}_{\mathrm{diag}}}
\left[
\ell_{\mathrm{pose}}
\left(
\widehat{\boldsymbol{\xi}}_{t,\ell,s},
\boldsymbol{\xi}_{t}
\right)
\right]
\end{equation}
denote the held-out probe error, and let $e_{\mathrm{chance}}$ denote the error
of a mean-motion predictor. 
We convert the error into a higher-is-better decodability score:
\begin{equation}
D_{\ell,s}
=
\left[
1-
\frac{e_{\ell,s}}
{e_{\mathrm{chance}}+\epsilon}
\right]_{[0,1]},
\qquad
[x]_{[0,1]}
=
\min\{1,\max\{0,x\}\},
\end{equation}
where $\epsilon$ is a small numerical constant. 
The raw values of $D_{\ell,s}$ are retained for reporting; normalization is
applied only when constructing the target-selection profiles.

\paragraph{Causal intervention.}
At each location $c=(\ell,s)$, we keep the prompt, initial noise, and random
seed fixed. 
Let $a\in\mathcal{A}$ denote one of four interventions:
\emph{Mask}, \emph{Shuffle}, \emph{Replace}, and \emph{Inject}. 
Mask removes the temporal state, Shuffle permutes its temporal order, Replace
substitutes the corresponding state from another clip, and Inject inserts a
pose-matched donor state.

For each target clip, donor clips are sampled from other clips in
$\mathcal{D}_{\mathrm{diag}}$ with the same frame length and valid pose
estimates. 
For Inject, the donor is selected according to the closest relative-pose
distance; for Replace, the donor is sampled without pose matching. 
Let $\mathbf{h}_c$ be the original hidden state and let
$\Delta\mathbf{h}_c^{(a)}$ be the change induced by intervention $a$. 
The relative intervention magnitude is
\begin{equation}
r_c^{(a)}
=
\frac{
\left\|
\Delta\mathbf{h}_c^{(a)}
\right\|_2
}{
\left\|
\mathbf{h}_c
\right\|_2+\epsilon
}.
\end{equation}
The random control is constructed from the same donor pool and rescaled to
match the intervention magnitude:
\begin{equation}
\Delta\mathbf{h}_c^{(\mathrm{rand},a)}
=
\frac{
\left\|
\Delta\mathbf{h}_c^{(a)}
\right\|_2
}{
\left\|
\Delta\mathbf{h}_c^{(\mathrm{donor})}
\right\|_2+\epsilon
}
\Delta\mathbf{h}_c^{(\mathrm{donor})}.
\end{equation}

Let $Y_c^{(a)}$ and
$Y_c^{(\mathrm{rand},a)}$ denote the videos generated under the intervention
and its norm-matched random control. 
The random-control-adjusted causal effect is
\begin{equation}
C_c^{(a)}
=
\eta_a
\left[
\mathbb{E}
\left[
E_{\mathrm{traj}}\!\left(Y_c^{(a)}\right)
\right]
-
\mathbb{E}
\left[
E_{\mathrm{traj}}\!\left(Y_c^{(\mathrm{rand},a)}\right)
\right]
\right],
\end{equation}
where $E_{\mathrm{traj}}(Y;d)$ measures the trajectory error of the
generated video $Y$ against the ground-truth camera trajectory
$\mathbf{T}^{\mathrm{ref},d}_{1:T}$ associated with diagnostic clip $d$.
This trajectory is an evaluation reference rather than a generation
condition. 
We set $\eta_a=+1$ for disruptive interventions and $\eta_a=-1$ for Inject, so
that a larger value always indicates a stronger causal effect. 
The aggregate causal score is
\begin{equation}
C_{\ell,s}
=
\frac{1}{|\mathcal{A}|}
\sum_{a\in\mathcal{A}} C_{(\ell,s)}^{(a)}.
\end{equation}

For $A\in\{L,S\}$, we average over the complementary axis:
\begin{equation}
D_i^A
=
\operatorname{Avg}_{j}D_{i,j},
\qquad
C_i^A
=
\operatorname{Avg}_{j}C_{i,j}.
\end{equation}
Each profile is independently min--max normalized:
\begin{equation}
\mathcal{N}(x_i)
=
\frac{x_i-\min_j x_j}
{\max_j x_j-\min_j x_j+\epsilon}.
\end{equation}
The joint profile used for target selection is
\begin{equation}
J_i^A
=
\frac{1}{2}
\left[
\mathcal{N}(D_i^A)
+
\mathcal{N}(C_i^A)
\right].
\end{equation}
Let $q_i$ denote the actual layer coordinate or the reindexed
denoising-progress coordinate $s$. For timestep profiles, the coordinates
are ordered by increasing $s$, from earlier to later denoising states. The
positive decline rate is
\begin{equation}
g_i^A
=
\left[
\frac{J_i^A-J_{i+1}^A}
{q_{i+1}-q_i}
\right]_+,
\qquad
[x]_+=\max\{0,x\}.
\end{equation}
The resulting target sets are
\[
\mathcal{L}_{\mathrm{tar}}=\{10,12,14,20\},
\qquad
\mathcal{S}_{\mathrm{tar}}=\{398,799\}.
\]
The timestep set $\mathcal{S}_{\mathrm{tar}}$ is expressed in the reindexed
progress coordinate $s$, so $s=398$ is earlier than $s=799$ in the denoising
process. Layer 14 is the primary target, while layers
$\{10,12,20\}$ are auxiliary targets.
The alignment weights are proportional to the positive temporal decline rates:
\begin{equation}
\omega_{\ell,s}
=
\frac{g_s^S}
{\displaystyle\sum_{s'\in\mathcal{S}_{\mathrm{tar}}}g_{s'}^S+\epsilon},
\qquad
\ell\in\mathcal{L}_{\mathrm{tar}}.
\end{equation}

We report the raw $D_{\ell,s}$ and $C_{\ell,s}$ values, the intervention
magnitudes $r_c^{(a)}$, and their uncertainty estimates. 
Specifically, we bootstrap diagnostic clips~\citep{efron1979bootstrap} while preserving all intervention
variants belonging to the same clip. 
For a quantity $q$, the $95\%$ confidence interval is computed from the
bootstrap distribution as
\begin{equation}
\operatorname{CI}_{95}(q)
=
\left[
Q_{0.025}(q^{(1:B)}),
Q_{0.975}(q^{(1:B)})
\right],
\end{equation}
where $B$ is the number of bootstrap replicates and $Q_\alpha$ denotes the
$\alpha$-quantile.

\subsection{Additional Details of Delta-Token Alignment}
\label{app:delta_token_details}

The frozen streaming 3D foundation model~\citep{chen2026lingbotmap} produces geometric states
\begin{equation}
\mathbf{z}_t^{\mathrm{T}}
=
\Phi^{\mathrm{T}}(x_{1:t}),
\end{equation}
where $x_{1:t}$ denotes the observed video stream up to frame $t$. 
The teacher delta token is encoded from two consecutive geometric states:
\begin{equation}
\boldsymbol{\delta}_t^{\mathrm{T}}
=
E_{\Delta}
\left(
\mathbf{z}_t^{\mathrm{T}},
\mathbf{z}_{t+1}^{\mathrm{T}}
\right).
\end{equation}
The teacher delta-token modules are trained with transition reconstruction,
temporal prediction, and relative-pose supervision:
\begin{equation}
\mathcal{L}_{\mathrm{T}}
=
\mathcal{L}_{\mathrm{tr}}
+
\lambda_{\mathrm{temp}}\mathcal{L}_{\mathrm{temp}}
+
\lambda_{\mathrm{pose}}^{T}\mathcal{L}_{\mathrm{pose}}^{T}.
\end{equation}
Here, $\mathcal{L}_{\mathrm{tr}}$ reconstructs the next teacher state,
$\mathcal{L}_{\mathrm{temp}}$ predicts the next delta token, and
$\mathcal{L}_{\mathrm{pose}}^{T}$ supervises the relative camera motion
$\Delta\mathbf{T}_t=\mathbf{T}_t^{-1}\mathbf{T}_{t+1}$. 
After this stage, the teacher encoder, delta encoder, transition decoder,
temporal predictor, and camera decoder are frozen.

For a selected VGM layer $\ell$, the VGM transition feature is mapped into the
teacher delta-token space by
\begin{equation}
\boldsymbol{\delta}_{t,\ell,s}^{\mathrm{V}}
=
Q_{\ell}
\left(
\mathbf{u}_{t,\ell,s}
\right),
\end{equation}
where $Q_{\ell}$ is a layer-specific projector. 
With the VGM and teacher modules frozen, the projectors are trained using
\begin{equation}
\mathcal{L}_{\mathrm{map}}
=
\sum_{\ell\in\mathcal{L}_{\mathrm{tar}}}
\sum_{s\in\mathcal{S}_{\mathrm{tar}}}
\omega_{\ell,s}
\mathbb{E}_t
\left[
\left\|
\boldsymbol{\delta}_{t,\ell,s}^{\mathrm{V}}
-
\boldsymbol{\delta}_{t}^{\mathrm{T}}
\right\|_2^2
\right].
\end{equation}
The projectors are frozen after training. 
In the final stage, LoRA adapters~\citep{hu2022lora} are updated only at
$\mathcal{L}_{\mathrm{tar}}$ and $\mathcal{S}_{\mathrm{tar}}$ using the
pose-supervision and delta-token matching losses defined in
Sec.~3.3. 
The teacher modules and projectors are used only during training and introduce
no computation during inference.

\subsection{More Evidence for geometric state Forgetting}
\label{app:more_forgetting_evidence}

\begin{figure*}[t]
    \centering
    \includegraphics[
        width=\textwidth,
        keepaspectratio
    ]{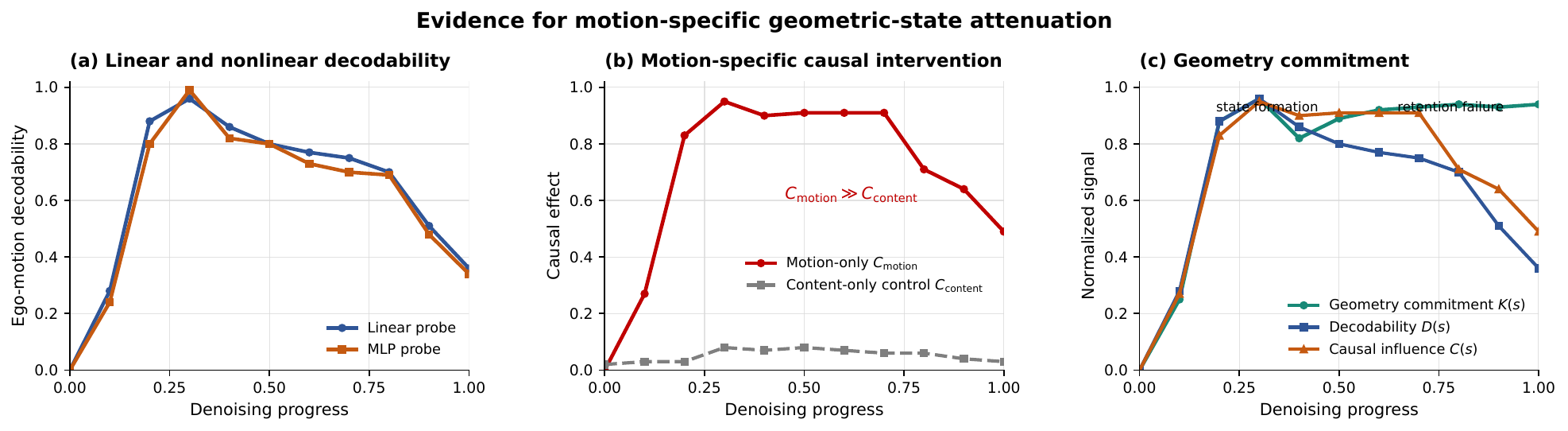}
    \caption{
    \textbf{Additional evidence for geometric state forgetting.}
    (a) Linear and MLP probes both show a rise-and-fall pattern in
    ego-motion decodability.
    (b) Motion-only interventions produce substantially larger causal
    effects than norm-matched content-only controls.
    (c) Geometry commitment $K(s)$ rises early and remains high, while
    the decodability $D(s)$ and motion-specific causal effect
    $C(s)$ reused from (a) and (b), respectively, decline during late
    denoising.
    Both probes and causal effects decline at later stages, while geometry commitment remains high, supporting the existence of geometric state forgetting.
    }

    \label{fig:forgetting_evidence}
\end{figure*}

The original decodability and intervention analyses may admit several
alternative explanations. A late decrease in linear decodability could
result from nonlinear re-encoding, while whole-state interventions may
change appearance and scene content together with ego-motion. Moreover,
camera geometry may already have been committed to the evolving video
latent before the measured hidden-state signals decline, as suggested by
the early emergence of camera motion reported in AC3D~\citep{bahmani2025ac3d}.
We therefore conduct three complementary analyses. The results are
summarized in Fig.~\ref{fig:forgetting_evidence}.

\paragraph{Linear and nonlinear decodability.}
We evaluate ego-motion decodability using both a linear probe and a
small MLP probe. As shown in Fig.~\ref{fig:forgetting_evidence}(a), both
probes exhibit a consistent rise-and-fall pattern over denoising.
Thus, the late-stage decrease is not specific to the linear readout
used in the main diagnosis. The result supports reduced accessibility
of the ego-motion state under both tested probe families, while making
no assumption that all possible nonlinear representations are
unrecoverable.

\paragraph{Motion-specific causal intervention.}
To isolate the causal role of ego-motion from appearance and scene
content, we compare an intervention applied to the ego-motion-related
delta-token component with a norm-matched intervention applied to
non-motion content. Let $C_{\mathrm{motion}}(s)$ denote the change in
the generated camera trajectory caused by the motion-only intervention,
and let $C_{\mathrm{content}}(s)$ denote the corresponding effect of
the content control. Figure~\ref{fig:forgetting_evidence}(b) shows that
\[
C_{\mathrm{motion}}(s) \gg C_{\mathrm{content}}(s)
\]
throughout the main diagnostic range, indicating that the measured
intervention effect is primarily attributable to the motion-specific
component rather than to a generic perturbation of the hidden state.
The motion-specific causal effect also decreases during late denoising.

\paragraph{Geometry commitment and late attenuation.}
Finally, we compare the geometry-commitment score $K(s)$ of the evolving
video latent with the internal ego-motion signals. Here, $D(s)$ is the
same linear-probe decodability curve reported in
Fig.~\ref{fig:forgetting_evidence}(a), and $C(s)$ is the same
motion-only causal-effect curve reported in
Fig.~\ref{fig:forgetting_evidence}(b). This shared construction allows
the three quantities to be compared directly. As shown in
Fig.~\ref{fig:forgetting_evidence}(c), $K(s)$ rises rapidly during early
denoising and remains high thereafter, whereas both $D(s)$ and $C(s)$
decline at later stages. This temporal separation indicates that the
geometric state has already formed in the evolving latent when the
measured hidden-state accessibility and causal effect begin to
attenuate. It therefore supports our interpretation of the late-stage
decline as a retention failure of the internal ego-motion state, rather
than a failure to form camera geometry in the first place.

\section{Extended Evaluation of CausalAlign}
\label{app:extended_evaluation}

This section first clarifies the metric definitions and training settings used in
the I2V evaluation in the main paper.  
We then explain why the unconstrained I2V protocol does not use direct ground-truth trajectory error. 
Finally, we report two complementary experiments: text-to-video generation and camera-trajectory-conditioned generation.

\subsection{Metric Definitions and Normalization}
\label{app:metric_definitions}

For generated clip $i$, let $\mathbf{p}_{i,t}$ and $\mathbf{R}_{i,t}$ denote the
estimated camera position and rotation at frame $t$. 
Let $\Delta\tau$ be the frame interval and let $d_{i,t}(u)$ be the estimated
depth at pixel $u$ and frame $t$. 
The clip-level median scene depth is
\begin{equation}
d_i
=
\operatorname{median}_{t,u} d_{i,t}(u).
\end{equation}
The depth-normalized translational speed and angular speed are
\begin{equation}
v_{i,t}
=
\frac{
\left\|
\mathbf{p}_{i,t+1}-\mathbf{p}_{i,t}
\right\|_2
}{
d_i\Delta\tau
},
\qquad
\omega_{i,t}
=
\frac{
\left\|
\log
\left(
\mathbf{R}_{i,t}^{\top}\mathbf{R}_{i,t+1}
\right)
\right\|_2
}{
\Delta\tau
}.
\end{equation}
For $m\in\{v,\omega\}$, let
\begin{equation}
\overline{m}_i
=
\frac{1}{T-1}
\sum_{t=1}^{T-1}m_{i,t}
\end{equation}
denote the clip-level mean. 
Let $(\mu_m^{\mathrm{real}},\sigma_m^{\mathrm{real}})$ denote the mean and
standard deviation of metric $m$ on the real-video validation distribution. 
We define the proximity score
\begin{equation}
S_m(i)
=
\exp
\left(
-\frac{
\left|
\overline{m}_i-\mu_m^{\mathrm{real}}
\right|
}{
\sigma_m^{\mathrm{real}}+\epsilon
}
\right),
\end{equation}
and report its dataset-level average:
\begin{equation}
M_m
=
\frac{1}{N}
\sum_{i=1}^{N}S_m(i).
\end{equation}
Thus, ``Mean Trans. Prox.'' and ``Mean Angular Prox.'' correspond to
$M_v$ and $M_\omega$, respectively. These scores measure proximity to
the real-data mean of the corresponding motion magnitude. A value
closer to one indicates that the generated clip-level mean is closer
to the real-data mean; it does not by itself measure agreement with the
full motion distribution.

For the composite gain used in the update-placement experiments, let $m_{i,k}$
denote the value of metric $k$ for clip $i$. 
Let $a_k$ and $b_k$ be fixed normalization bounds computed over the comparison
set. 
The direction-corrected normalized score is
\begin{equation}
q_{i,k}
=
\begin{cases}
\left[
\dfrac{m_{i,k}-a_k}{b_k-a_k+\epsilon}
\right]_{[0,1]},
& k\in\mathcal{K}_{\uparrow},\\[1.2ex]
\left[
\dfrac{b_k-m_{i,k}}{b_k-a_k+\epsilon}
\right]_{[0,1]},
& k\in\mathcal{K}_{\downarrow},
\end{cases}
\end{equation}
where $[x]_{[0,1]}=\min\{1,\max\{0,x\}\}$. 
The dataset-level score for method $M$ is
\begin{equation}
q_k^{(M)}
=
\frac{1}{N}
\sum_{i=1}^{N}q_{i,k}^{(M)}.
\end{equation}
The composite gain of an adapted model over the base model is
\begin{equation}
G
=
100
\left[
\sum_k w_k q_k^{(\mathrm{adapted})}
-
\sum_k w_k q_k^{(\mathrm{base})}
\right],
\qquad
\sum_k w_k=1.
\end{equation}
For the update-placement study, the composite contains rotation RPE,
translation-direction RPE, Sim(3) ATE, endpoint error, speed-profile error, and
depth AbsRel.

\subsection{Training and Inference Details}
\label{app:training_details}

We instantiate CausalAlign on WAN2.2-14B-I2V~\citep{wan2025} using DL3DV~\citep{ling2024dl3dv} and RealEstate10K~\citep{zhou2018stereo}.
Each dataset is divided into $90\%$ training data and $10\%$ held-out evaluation
data. 
Following the three-stage procedure in Fig.~\ref{fig:method}, the three stages
use fixed $30\%$, $10\%$, and $100\%$ subsets of the training split,
respectively.

In Stage 1, the teacher delta-token modules, including the delta encoder,
transition decoder, temporal predictor, and camera decoder, are trained with
\begin{equation}
\lambda_{\mathrm{temp}}=0.1,
\qquad
\lambda_{\mathrm{pose}}^{T}=0.2.
\end{equation}
In Stage 2, the VGM-side projectors are trained with
$\mathcal{L}_{\mathrm{map}}$ while the VGM and all teacher modules remain frozen.
In Stage 3, LoRA adapters are optimized only at
\begin{equation}
\mathcal{L}_{\mathrm{tar}}=\{10,12,14,20\},
\qquad
\mathcal{S}_{\mathrm{tar}}=\{398,799\},
\end{equation}
where layer $14$ is the primary target and layers $\{10,12,20\}$ are auxiliary
targets. 
The final objective is
\begin{equation}
\mathcal{L}_{\mathrm{pose}}
+
\lambda_{\Delta}\mathcal{L}_{\Delta},
\qquad
\lambda_{\Delta}=0.2.
\end{equation}
The original VGM backbone, LingBot-Map, teacher modules, and VGM-side projectors
remain frozen during the final stage. 
Only the adapted VGM is retained at inference, and no teacher or external
geometric cache is used.

\paragraph{Why the main I2V protocol does not use direct ground-truth pose error.}
In the main first-frame-and-prompt I2V protocol, the model receives only the
first frame and a text prompt. The future video is therefore one-to-many:
multiple camera trajectories can be valid continuations of the same input.
Directly comparing an unconstrained generated trajectory with a single
ground-truth trajectory can consequently penalize valid alternative motions.

This differs from the causal diagnosis in Sec.~\ref{sec:causal_targeting}.
There, the ground-truth trajectory of the corresponding diagnostic video is
used only as a fixed measurement reference for
$E_{\mathrm{traj}}(Y;d)$. It is never provided to the generator. The
diagnostic score measures how an intervention changes the generated trajectory
error relative to a norm-matched random control, thereby measuring the
sensitivity of the generation process to the intervened internal state. It
does not require the generated video to reproduce the reference trajectory.

For the main unconstrained I2V evaluation, methods whose native inference does
not require an explicit camera trajectory receive only the first frame and
the text prompt. These methods are evaluated with matched output length,
resolution, frame rate, sampling budget, and random seeds. For methods whose
standard inference requires an explicit camera trajectory and a 3D cache, such
as GEN3C and Voyager~\citep{ren2025gen3c,huang2025voyager}, we follow their
native inference interfaces: the ground-truth camera trajectory is provided as
the camera-control condition, and a fresh 3D cache is constructed from the
available input frame(s), without using future video frames. These methods are
reported as trajectory-conditioned baselines rather than as unconstrained
generators.

Regardless of the input interface, every generated video is processed by the
same frozen, independent reconstruction pipeline. For the main intrinsic
comparison, trajectory analysis is based only on the reconstructed video and
uses motion magnitude, geometric self-consistency, temporal dynamics, and
visual quality. Neither the conditioning trajectory nor its ground-truth
trajectory is used as the target for these intrinsic metrics. Thus, the
trajectory-conditioned baselines are not evaluated by a ground-truth pose
matching loss, while the evaluator and intrinsic metrics remain shared across
methods. Results from unconstrained and trajectory-conditioned protocols are
therefore interpreted separately.

To complement this intrinsic evaluation, we introduce a separate
camera-trajectory-conditioned protocol. In this setting, the reference camera
trajectory $\mathbf{T}_{1:T}^{*}$ is provided as an explicit condition, and
direct comparison with the target trajectory is well-defined. The results are
reported in Sec.~\ref{sec:extended_camera_control}.

We next evaluate CausalAlign in two additional settings. The first tests
whether the learned alignment transfers to text-to-video generation. The
second evaluates direct trajectory accuracy under explicit camera conditioning.
Within each setting, methods share evaluation inputs, output length,
resolution, frame rate, sampling budgets, and evaluation procedures.

\subsection{Text-to-Video Generation}
\label{sec:extended_t2v}

We adapt Wan2.2-14B-T2V with the same training hyperparameters and evaluate it on
the same test split as the main I2V experiment. 
Text prompts are generated from RealEstate10K clips~\citep{zhou2018stereo} following the captioning
procedure used by CameraCtrl~\citep{he2024cameractrl}. 
Both the native and adapted models receive only these prompts.

\paragraph{Visual quality.}
Table~\ref{tab:t2v_visual} reports VBench quality, semantic, and smoothness
scores~\citep{huang2024vbench}, together with FVD~\citep{unterthiner2018towards}, CLIP~\citep{radford2021learning}, SSIM~\citep{wang2004image}, and LPIPS~\citep{zhang2018unreasonable}. 
CausalAlign improves quality, semantic score, FVD, and LPIPS, matches
smoothness, and has lower CLIP and SSIM scores.

\begin{table}[!htbp]
\centering\small

\caption{
\textbf{Visual quality in text-to-video generation.}
We compare Wan2.2-14B-T2V and CausalAlign using VBench and complementary visual metrics.
CausalAlign improves quality, semantic score, FVD, and LPIPS while preserving smoothness, with slightly lower CLIP and SSIM scores.
}

\label{tab:t2v_visual}
\setlength{\tabcolsep}{3pt}
\begin{tabular*}{\linewidth}{@{\extracolsep{\fill}}lrrrrrrr@{}}
\toprule
& \multicolumn{3}{c}{VBench} & \multicolumn{4}{c}{Other visual metrics} \\
\cmidrule(lr){2-4}\cmidrule(lr){5-8}
Method & Quality$\uparrow$ & Semantic$\uparrow$ & Smooth.$\uparrow$ & FVD$\downarrow$ & CLIP$\uparrow$ & SSIM$\uparrow$ & LPIPS$\downarrow$ \\
\midrule
Wan2.2-14B-T2V & 0.850 & 0.729 & \textbf{0.979} & 183.6 & \textbf{0.331} & \textbf{0.728} & 0.181 \\
CausalAlign & \textbf{0.852} & \textbf{0.730} & \textbf{0.979} & \textbf{183.4} & 0.320 & 0.726 & \textbf{0.175} \\
\bottomrule
\end{tabular*}
\end{table}

\paragraph{Intrinsic ego motion.}
The frozen pose, depth, optical-flow, and feature estimators from the main
evaluation process all generated videos. 
Motion magnitudes are compared with the real-video reference distribution,
whereas geometric consistency and temporal dynamics measure internal video
consistency. 
Table~\ref{tab:t2v_ego_motion} shows improved motion magnitudes, reprojection,
depth--flow, and epipolar errors, together with higher smoothness and
lower jerk.

\begin{table}[!htbp]
\centering\small
\caption{
\textbf{Ego-motion consistency in text-to-video generation.}
We evaluate Wan2.2-14B-T2V and CausalAlign using the same text prompts and geometric estimators.
CausalAlign improves all seven metrics across motion magnitude, geometric consistency, and temporal dynamics.
}

\label{tab:t2v_ego_motion}
\setlength{\tabcolsep}{3pt}
\begin{tabular*}{\linewidth}{@{\extracolsep{\fill}}lrrrrrrr@{}}
\toprule
& \multicolumn{2}{c}{Motion magnitude} & \multicolumn{3}{c}{Geometric consistency} & \multicolumn{2}{c}{Temporal dynamics} \\
\cmidrule(lr){2-3}\cmidrule(lr){4-6}\cmidrule(lr){7-8}
Method & \shortstack{Mean Trans. \\Prox.$\uparrow$} & \shortstack{Mean Angular\\ Prox.$\uparrow$} & \shortstack{Reproj.\\(px)$\downarrow$} & \shortstack{Depth--flow\\(px)$\downarrow$} & \shortstack{Epipolar\\(px)$\downarrow$} & Smooth.$\uparrow$ & Jerk$\downarrow$ \\
\midrule
Wan2.2-14B-T2V & 0.63 & 0.68 & 8.2 & 7.1 & 5.4 & 0.78 & 0.181 \\
CausalAlign & \textbf{0.78} & \textbf{0.90} & \textbf{6.9} & \textbf{5.3} & \textbf{3.4} & \textbf{0.86} & \textbf{0.154} \\
\bottomrule
\end{tabular*}
\end{table}

\subsection{Camera-Trajectory-Conditioned Generation}
\label{sec:extended_camera_control}

We apply CausalAlign to Wan2.2-Fun-A14B-Control-Camera~\citep{alibabapai2025wanfun_camera} and compare it with the
native model and FantasyWorld~\citep{dai2026fantasyworld}. 
Each method receives the same first frame, ground-truth camera trajectory, and
text prompt. 
Generated trajectories are recovered with a shared frozen pose estimator and
compared with the conditioning trajectories using identical temporal
correspondence, coordinate conventions, and trajectory alignment.

\begingroup\emergencystretch=1.5em
Table~\ref{tab:camera_control} reports relative rotation and translation errors
(RRE~\citep{zhang2018tutorial} and RTE~\citep{geiger2012we}), absolute trajectory error (ATE~\citep{sturm2012benchmark}), final displacement error (FDE~\citep{alahi2016social}),
drift~\citep{geiger2012we}, smoothness~\citep{hogan2009sensitivity}, and jerk~\citep{flash1985coordination}. 
CausalAlign gives the best values on six of the seven metrics; FantasyWorld has the lowest RTE.
\par\endgroup

\subsection{More Visualization For Mobile Manipulation Rollouts}
\label{sec:wam_vis}
\begin{figure}[t]
    \centering
    \includegraphics[width=\linewidth]{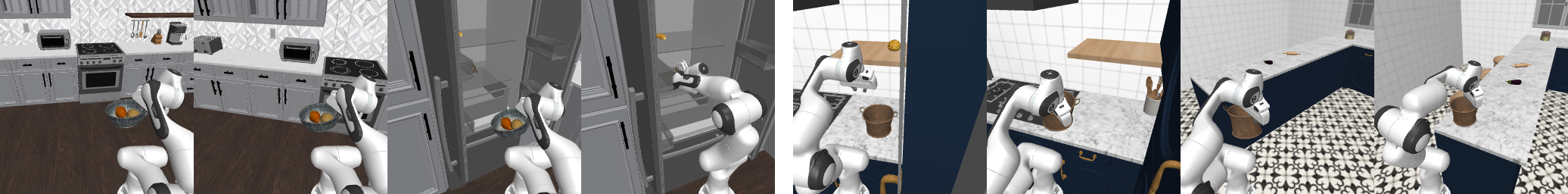}

\caption{
\textbf{Successful mobile manipulation rollouts.}
We visualize two policy rollouts on RoboCasa365, showing a robot placing a bowl into a refrigerator and moving a bread basket to a counter.
Both examples demonstrate successful execution of mobile manipulation tasks.
}

    \label{fig:wam_mobile_rollouts}
\end{figure}

Here we give two more successful examples on RoboCasa365~\citep{nasiriany2026robocasa365} in Fig.~\ref{fig:wam_mobile_rollouts}. Meanwhile, we provide our experiment setting in Table~\ref{tab:cadet_protocol}.

\begin{table}[!htbp]
\centering\small

\caption{
\textbf{Camera trajectory accuracy under explicit conditioning.}
We compare CausalAlign with the native Wan2.2 camera control model and FantasyWorld using identical camera trajectories, initial frames, and text prompts.
CausalAlign achieves the best results on six of seven metrics, while FantasyWorld has the lowest relative translation error.
}

\label{tab:camera_control}
\setlength{\tabcolsep}{3pt}
\begin{tabular*}{\linewidth}{@{\extracolsep{\fill}}lrrrrrrr@{}}
\toprule
& \multicolumn{2}{c}{Relative pose} & \multicolumn{3}{c}{Global trajectory} & \multicolumn{2}{c}{Temporal dynamics} \\
\cmidrule(lr){2-3}\cmidrule(lr){4-6}\cmidrule(lr){7-8}
Method & \shortstack{RRE\\($^\circ$)$\downarrow$} & \shortstack{RTE\\(\%)$\downarrow$} & \shortstack{ATE\\(m)$\downarrow$} & \shortstack{FDE\\(m)$\downarrow$} & \shortstack{Drift\\(\%)$\downarrow$} & Smooth.$\uparrow$ & Jerk$\downarrow$ \\
\midrule
Native & 3.82 & 7.70 & 0.214 & 0.263 & 13.6 & 0.69 & 0.189 \\
FantasyWorld & 3.72 & \textbf{6.60} & 0.169 & 0.208 & 10.5 & 0.82 & 0.159 \\
CausalAlign & \textbf{3.21} & 6.80 & \textbf{0.167} & \textbf{0.201} & \textbf{10.2} & \textbf{0.85} & \textbf{0.155} \\
\bottomrule
\end{tabular*}
\end{table}

\section{Additional Ablation Studies}
\label{sec:cadet_extended_ablations}

\subsection{Structured Delta-Space Representation}
\label{sec:cadet_partition_protocol}

\paragraph{Representations and Stage-1 supervision.}
Table~\ref{tab:cadet_core} evaluates the frozen teacher delta-token
representation learned in Stage 1. The structured encoder produces
\begin{equation}
\boldsymbol{\delta}_{t}^{\mathrm{T}}
=
\left[
\boldsymbol{\delta}_{t}^{\mathrm{ego}};
\boldsymbol{\delta}_{t}^{\mathrm{geo}};
\boldsymbol{\delta}_{t}^{\mathrm{vis}}
\right],
\end{equation}
where the three regions encode relative camera motion, camera-local
geometric changes such as depth and surface-normal differences, and RGB
changes, respectively.

These regions are structured target channels rather than three additional
losses. The camera decoder uses the pose target to provide explicit
ego-motion supervision through $\mathcal{L}_{\mathrm{pose}}^{T}$. The
transition decoder reconstructs the next teacher state, so
$\mathcal{L}_{\mathrm{tr}}$ supervises the transition-related geometric and
visual content in that state. The temporal predictor forecasts the complete
next delta token, and $\mathcal{L}_{\mathrm{temp}}$ is therefore applied to
the concatenated token, including all three regions. Thus, the pose,
depth/normal, and RGB targets jointly implement the Stage-1 objective in
Eq.~(7), rather than defining separate objectives.

For the unpartitioned controls, the pair and difference encoders also
produce 128-dimensional tokens and are trained with the same three loss
terms and the same pose, geometric, and visual targets, but without
explicitly separating the token into semantic regions. All encoders receive
40,000 updates: 10,000 on 4,000 training scenes and 30,000 on 8,312 scenes.
Camera poses and target geometry provide the supervision labels.

\paragraph{Readout protocol.}
Independent two-layer, width-256 probes are trained for 3,000 updates with batch size 128 on 4,000 scenes. Learning rates are selected from $\{10^{-5},10^{-4},10^{-3}\}$ using 150-step validation runs. Evaluation covers 256 development scenes and three seeds. Each scene contributes nine frame pairs, combining source indices $\{0,4,8\}$ with gaps $\{1,2,4\}$ in a 21-frame, stride-4 feature sequence. Scores are averaged over frame pairs within each scene, then over scenes and seeds.

\paragraph{Targets and metrics.}
Ego-motion readouts are evaluated against DL3DV poses~\citep{ling2024dl3dv} using rotation and translation-direction angular errors. Geometry probes predict log-depth changes on a $16\times28$ grid, which are added to the common source log-depth and compared with frozen LingBot-Map estimates~\citep{chen2026lingbotmap} on shared valid pixels. We report $\mathrm{AbsRel}=\langle|\hat d-d|/d\rangle$ and $\delta_1=\langle\mathbf{1}[\max(\hat d/d,d/\hat d)<1.25]\rangle$~\citep{eigen2014depth}. Visual probes predict RGB changes on a $32\times56$ grid; reconstruction from the common source image is evaluated with SSIM~\citep{wang2004image} and AlexNet LPIPS~\citep{zhang2018unreasonable}. Copying the source image gives SSIM $0.29397$ and LPIPS $0.08924$, versus $0.30283$ and $0.08676$ for the complete delta token.

\paragraph{Partition specialization.}
For the matched-width comparisons in Table~\ref{tab:cadet_core}(a), PCA~\citep{jolliffe2016principal} fitted on training data projects the supervised unpartitioned encoders to 64, 40, and 24 dimensions. Delta regions lead all 12 paired metric comparisons. Table~\ref{tab:cadet_crossprobe} further evaluates every region on every task at a common width of 24 dimensions, projecting the ego and geometry regions with training-only PCA. Each region performs best on its designated task.

\begin{table}[!htbp]
\centering\small

\caption{
\textbf{Specialization of delta-token regions.}
We evaluate the ego-motion, geometry, and visual regions using probes with a common dimension of 24 on 256 scenes across three seeds.
Each region performs best on its corresponding task, demonstrating the effectiveness of semantic partitioning.
}

\label{tab:cadet_crossprobe}
\setlength{\tabcolsep}{4pt}
\begin{tabular*}{\linewidth}{@{\extracolsep{\fill}}lrrrrrr@{}}
\toprule
& \multicolumn{2}{c}{Ego motion} & \multicolumn{2}{c}{Geometry} & \multicolumn{2}{c}{Visual} \\
\cmidrule(lr){2-3}\cmidrule(lr){4-5}\cmidrule(lr){6-7}
Region & Rotation ($^\circ$)$\downarrow$ & Direction ($^\circ$)$\downarrow$ & AbsRel$\downarrow$ & $\delta_1\uparrow$ & SSIM$\uparrow$ & LPIPS$\downarrow$ \\
\midrule
Ego & \textbf{2.429} & \textbf{7.558} & 0.3115 & 0.6764 & 0.2881 & 0.08964 \\
Geometry & 13.070 & 49.822 & \textbf{0.2292} & \textbf{0.7448} & 0.2908 & 0.08894 \\
Visual & 14.682 & 65.830 & 0.3188 & 0.6620 & \textbf{0.3050} & \textbf{0.08643} \\
\bottomrule
\end{tabular*}
\end{table}

\paragraph{Original features and extraction controls.}
The original LingBot-Map frame representation concatenates the final camera token and the mean of 777 patch tokens, yielding 4096 dimensions. The pooled pair A concatenates two frame vectors (8192D), while B takes their difference (4096D). Both retain their original width, with coordinate-wise standardization fitted on probe-training data. Table~\ref{tab:cadet_core}(b) compares these frozen teacher features with the complete learned delta representation; panel (a) isolates partitioning under matched supervision. Delta leads all six readout metrics with the original eight-frame streaming context, with whole-sequence extraction, and with whole-sequence extraction plus four-layer averaging. A full-spatial-grid difference provides a separate 133,120D control: its AbsRel $0.2109$, SSIM $0.3533$, and LPIPS $0.08575$ improve on the corresponding delta readouts.

\subsection{Update Placement and Temporal Modulation}
\label{sec:cadet_placement_evidence}

\paragraph{Independent adapter placement.}
We independently train a LoRA adapter for each of five layer sets in the
second transformer of Wan2.2-14B. All runs use the same training data,
objective, LoRA configuration, optimization budget, and inference
protocol, and the resulting updates are matched in effective global
norm. The fixed-source video-to-video evaluation uses 32 scenes, one
generation seed, ten inference steps, and identical sources, prompts,
scheduler, and initial noise. The composite combines rotation RPE, translation-direction RPE, Sim(3) ATE, endpoint error, speed-profile error, and depth AbsRel, using frozen LingBot-Map estimates from source videos as references. Positive values denote improvements; 95\% confidence intervals use 50,000 scene-bootstrap resamples~\citep{efron1979bootstrap}.

Table~\ref{tab:cadet_placement_complete}(a) reports the complete placement comparison. The independently
trained adapter at layers 10, 12, 14, and 20 gives the largest gain
over the native model (3.109\%). Its gain over a norm-matched random
update is 2.224\%. This matched-adapter comparison isolates the effect
of target-layer placement.

\paragraph{Temporal modulation.}
Using 32 scenes and three seeds, we compare timestep-conditioned, static, time-averaged, and ensemble updates under the same fixed-source protocol. Table~\ref{tab:cadet_placement_complete}(b) reports all three contrasts. The ensemble has the largest mean gain over the static update; each interval includes zero.

\begin{table}[!htbp]
\centering\small

\caption{
\textbf{Effects of alignment placement and temporal modulation.}
We compare LoRA updates at different layer groups and strategies for temporal modulation on Wan2.2-14B, reporting composite gains with 95\% confidence intervals.
The layers selected by causal diagnosis achieve the largest gain of 3.109\%, while the temporal modulation comparisons have confidence intervals that include zero.
}

\label{tab:cadet_placement_complete}
\setlength{\tabcolsep}{3pt}
\begin{tabular*}{\linewidth}{@{\extracolsep{\fill}}lrrrr@{}}
\toprule
\multicolumn{5}{l}{\textit{(a) Layer placement}} \\
& \multicolumn{2}{c}{Versus native} & \multicolumn{2}{c}{Versus random update} \\
\cmidrule(lr){2-3}\cmidrule(lr){4-5}
Layer set & Gain & 95\% CI & Gain & 95\% CI \\
\midrule
10, 12, 14, 20 & +3.109 & [+0.808, +5.373] & +2.224 & [+0.302, +4.022] \\
16, 18, 20, 26 & +0.250 & [-2.483, +2.769] & -1.045 & [-4.391, +1.961] \\
20, 22, 24, 30 & -2.194 & [-5.552, +0.760] & -1.731 & [-5.315, +1.611] \\
10, 14, 20, 24 & -0.513 & [-3.595, +2.441] & -0.738 & [-4.136, +2.232] \\
10, 12, 14, 16, 18, 20 & -0.414 & [-2.919, +1.883] & +1.652 & [-0.348, +3.571] \\
\midrule
\multicolumn{5}{l}{\textit{(b) Temporal modulation}} \\
Comparison & Gain & 95\% CI & & \\
\midrule
Dynamic vs.\ static & +0.694 & [-0.949, +2.405] & & \\
Dynamic vs.\ time-averaged & -0.238 & [-2.055, +1.608] & & \\
Ensemble vs.\ static & +2.130 & [-0.057, +4.467] & & \\
\bottomrule
\end{tabular*}
\end{table}

\subsection{Auxiliary Objectives and Update Strength}
\label{sec:cadet_extra_motion_losses}

We compare local relative-pose supervision (RPE), kinematic supervision (speed, acceleration, and jerk), and their combination on Wan2.2-TI2V-5B. The three runs continue from the same parent checkpoint to the same endpoint, step 48. Local RPE gives the largest trajectory gain ($3.525\%$), with improvements concentrated in endpoint and motion-dynamics errors. Combining objectives reduces this gain to $1.299\%$ while improving geometry and temporal preservation (Table~\ref{tab:cadet_motion}).

In matched step-24 continuations, reducing the later-layer update scale from $0.50$ to $0.25$ increases trajectory gain from $1.750\%$ to $2.209\%$ and geometry gain from $-0.506\%$ to $0.761\%$. Temporal and visual gains decrease from $0.218\%$ and $0.059\%$ to $0.086\%$ and $0.021\%$, respectively. Both comparisons show that stronger or more numerous constraints need not improve all metric groups together.

\begin{table}[!htbp]
\centering\small

\caption{
\textbf{Effects of auxiliary motion objectives.}
We compare supervision using relative poses, motion dynamics, and their combination on Wan2.2-TI2V-5B under identical training settings.
Relative pose supervision achieves the largest trajectory gain of 3.525\%, while combining both objectives provides better geometric and temporal preservation.
}

\label{tab:cadet_motion}
\setlength{\tabcolsep}{5pt}
\begin{tabular*}{\linewidth}{@{\extracolsep{\fill}}lrrr@{}}
\toprule
Metric & Local RPE & Kinematic & RPE + kinematic \\
\midrule
\multicolumn{4}{l}{\textit{Local pose and global trajectory}} \\
Rotation RPE & +0.132 & +0.065 & +0.078 \\
Translation-direction RPE & +0.399 & +2.002 & +1.363 \\
Sim(3) ATE & +2.024 & +2.521 & +1.074 \\
Endpoint error & +5.602 & +4.952 & +3.089 \\
Path-ratio error & -0.031 & -0.048 & -0.024 \\
\midrule
\multicolumn{4}{l}{\textit{Motion dynamics}} \\
Speed-profile error & +6.629 & +3.536 & +2.035 \\
Acceleration error & +6.254 & +1.676 & +1.424 \\
Jerk error & +7.193 & +2.595 & +1.352 \\
\midrule
\multicolumn{4}{l}{\textit{Group summaries}} \\
Trajectory & +3.525 & +2.162 & +1.299 \\
Geometry & -0.468 & +0.475 & +0.643 \\
Temporal & -0.083 & -0.033 & +0.025 \\
Visual & +0.065 & +0.058 & +0.047 \\
\bottomrule
\end{tabular*}
\end{table}

\subsection{Additional 5B Configurations}
\label{sec:cadet_5b_configurations}

Tables~\ref{tab:cadet_5b_all} and~\ref{tab:cadet_5b_all_16} collect the 27 evaluated configurations, grouped by update design and training objective. They report trajectory, geometry, temporal, and visual improvements, their mean, and its 95\% confidence interval. These recipe comparisons include different continuation lengths; the matched controls are described above. Q/V and Q/K/V/O specify the self-attention projections being updated, and L22 denotes layer 22. Short configuration identifiers distinguish the remaining training recipes.

\begin{table}[!htbp]
\centering\small

\caption{
\textbf{Effects of adapter updates and weight averaging.}
We evaluate different update configurations and weight averaging strategies on the 5B backbone, reporting improvements over the original model.
Scaling the update at layer 22 to 0.25 achieves the highest mean gain of 0.769\% among the compared configurations.
}

\label{tab:cadet_5b_all}
\setlength{\tabcolsep}{3pt}
\begin{tabular*}{\linewidth}{@{\extracolsep{\fill}}lrrrrrr@{}}
\toprule
Configuration & Traj. & Geom. & Temp. & Visual & Mean & 95\% CI \\
\midrule
\multicolumn{7}{l}{\textit{Q/V updates and layer scaling}} \\
Q/V balanced & +2.223 & -0.609 & +0.153 & +0.036 & +0.451 & [-0.217, +1.090] \\
Q/V, L22 scale 0.25 & +2.209 & +0.761 & +0.086 & +0.021 & +0.769 & [+0.090, +1.454] \\
Q/V, L22 scale 0.50 & +1.750 & -0.506 & +0.218 & +0.059 & +0.380 & [-0.422, +1.174] \\
Q/V, long, unscaled & +0.292 & -0.378 & +0.150 & +0.045 & +0.027 & [-0.679, +0.721] \\
Attenuated L22 & +1.025 & -0.508 & +0.139 & +0.048 & +0.176 & [-0.611, +0.989] \\
Attenuated L22, long & +0.891 & +0.255 & +0.262 & +0.045 & +0.363 & [-0.248, +0.969] \\
\midrule
\multicolumn{7}{l}{\textit{Weight averaging}} \\
Q/V--L22 blend, 0.25 & +1.898 & -1.348 & +0.107 & +0.034 & +0.173 & [-0.547, +0.885] \\
Q/V--L22 blend, 0.50 & +2.615 & -0.624 & +0.241 & +0.062 & +0.574 & [-0.110, +1.193] \\
Q/V--L22 blend, 0.75 & +1.915 & -0.758 & +0.077 & +0.031 & +0.316 & [-0.497, +1.114] \\
\texttt{soup12\_75} & +1.420 & -0.342 & +0.143 & +0.077 & +0.324 & [-0.567, +1.224] \\
\bottomrule
\end{tabular*}
\end{table}

\begin{table}[!htbp]
\centering\small

\caption{
\textbf{Effects of training objectives and configurations.}
We compare different objectives for motion, geometry, and appearance under various training configurations on the 5B backbone.
Relative pose alignment at step 72 achieves the highest mean gain of 0.982\% among the compared configurations.
}

\label{tab:cadet_5b_all_16}
\setlength{\tabcolsep}{3pt}
\begin{tabular*}{\linewidth}{@{\extracolsep{\fill}}lrrrrrr@{}}
\toprule
Configuration & Traj. & Geom. & Temp. & Visual & Mean & 95\% CI \\
\midrule
\multicolumn{7}{l}{\textit{Motion objectives}} \\
Local RPE, step 48 & +3.525 & -0.468 & -0.083 & +0.065 & +0.760 & [-0.142, +1.626] \\
Kinematic, step 48 & +2.162 & +0.475 & -0.033 & +0.058 & +0.666 & [-0.078, +1.328] \\
RPE + kinematic, step 48 & +1.299 & +0.643 & +0.025 & +0.047 & +0.503 & [-0.148, +1.150] \\
Aligned RPE, step 72 & +1.728 & +1.987 & +0.172 & +0.040 & +0.982 & [+0.150, +1.796] \\
Aligned kinematic, step 72 & +0.249 & -1.373 & -0.008 & +0.042 & -0.273 & [-1.024, +0.489] \\
Aligned all, step 72 & +0.236 & +0.095 & +0.199 & +0.039 & +0.142 & [-0.486, +0.843] \\
\midrule
\multicolumn{7}{l}{\textit{Geometry, appearance, and other objectives}} \\
Geometry--visual, s12 & +1.763 & -0.430 & -0.059 & +0.053 & +0.332 & [-0.500, +1.149] \\
High-frequency, light & +1.020 & -0.277 & +0.071 & +0.042 & +0.214 & [-0.630, +0.982] \\
High-frequency, s8 & -0.209 & -0.519 & +0.074 & +0.079 & -0.144 & [-0.852, +0.525] \\
\texttt{current} & +0.951 & +0.137 & -0.277 & -0.115 & +0.174 & [-0.636, +0.967] \\
\texttt{gt\_full} & +1.310 & -0.390 & -0.200 & -0.049 & +0.168 & [-0.621, +0.928] \\
\texttt{gt\_half} & +0.089 & -0.804 & -0.205 & -0.037 & -0.239 & [-0.965, +0.473] \\
\texttt{gt\_preserve6k} & -0.774 & -1.392 & -0.538 & -0.087 & -0.698 & [-1.393, +0.033] \\
\texttt{dual\_short} & +0.941 & -0.553 & -0.184 & -0.001 & +0.051 & [-0.812, +0.913] \\
\texttt{dual\_flowmatch1} & +0.451 & -0.868 & -0.022 & -0.042 & -0.120 & [-0.909, +0.628] \\
\texttt{dual\_aux015} & +0.306 & -0.639 & -0.412 & -0.069 & -0.203 & [-1.018, +0.556] \\
Dual Q/K/V/O & -0.895 & -0.978 & -0.025 & +0.048 & -0.462 & [-1.318, +0.377] \\
\bottomrule
\end{tabular*}
\end{table}

\begin{table}[!htbp]
\centering\small
\caption{
\textbf{Training and evaluation settings for Fast-WAM.}
We summarize the training configuration and evaluation protocol for mobile manipulation on RoboCasa365 Target100.
The protocol evaluates the policy on 18 Atomic-Seen tasks with 50 rollouts per task, providing the settings used to assess downstream performance.
}

\label{tab:cadet_protocol}
\setlength{\tabcolsep}{4pt}
\begin{tabular*}{\linewidth}{@{\extracolsep{\fill}}lp{0.56\linewidth}@{}}
\toprule
Setting & Configuration \\
\midrule
Pretraining data & Human300, 100 demonstrations per task \\
Target demonstrations & 500 per task (Target100) \\
Evaluation scenes & Held-out target kitchens \\
Pretraining / evaluated fine-tuning step & 60k / 50k \\
Fine-tuning global batch & 128 \\
Fine-tuning optimizer & AdamW, weight decay 0.01, gradient clipping 1.0 \\
Fine-tuning learning rate & $3\times10^{-5}$, 3000-step warmup, cosine decay \\
Action / state dimension & 12 / 16 \\
Input cameras & 3  \\
Predicted action chunk / executed prefix & 32 / 8 \\
Denoising steps & 10 \\
Quantitative evaluation & 18 Atomic-Seen tasks, 50 rollouts per task \\
Evaluation seeds & 10000--10049 \\
Simulation renderer & CPU llvmpipe \\
\bottomrule
\end{tabular*}
\end{table}

\end{document}